\documentclass{article}
\usepackage{arxiv}

\usepackage[utf8]{inputenc} 
\usepackage[T1]{fontenc}    
\usepackage{hyperref}       
\usepackage{url}            
\usepackage{booktabs}       
\usepackage{amsfonts}       
\usepackage{nicefrac}       
\usepackage{microtype}      
\usepackage{xcolor}         
\usepackage{graphicx}
\usepackage{pifont}     
\usepackage{colortbl}   
\usepackage{float}
\usepackage{enumitem}
\usepackage{booktabs, multirow}
\usepackage[table]{xcolor}
\usepackage[normalem]{ulem}
\usepackage{amsmath}
\usepackage{tabularx}
\usepackage{booktabs}
\usepackage{subfig}
\usepackage{caption}
\usepackage{wrapfig}

\definecolor{l1color}{RGB}{156,179,210}
\definecolor{l2color}{RGB}{228,170,122}
\definecolor{l3color}{RGB}{172,120,132}

\newcommand{\cmark}{\textcolor{green!70!black}{\ding{51}}} 
\newcommand{\xmark}{\textcolor{red!80!black}{\ding{55}}}   
\newcommand{\first}[1]{\textbf{#1}$^{\ast}$}     
\newcommand{\second}[1]{\textbf{#1}$^{\dagger}$} 
\newcommand{\third}[1]{\textbf{#1}$^{\ddagger}$} 

\title{OmniTraffic: A Controllable Generation Pipeline and Benchmark for Spatio-Temporal Traffic Reasoning}

\author{
\textbf{Maonan Wang}$^{1,2}$,
\textbf{Zhengyan Huang}$^{2}$,
\textbf{Kemou Jiang}$^{3}$,
\textbf{Yuhang Fu}$^{3}$,
\textbf{Jiayue Zhu}$^{3}$,
\textbf{Yuxin Cai}$^{4}$, \\ 
\textbf{Xingchen Zou}$^{5}$,
\textbf{Qiaosheng Zhang}$^{2}$,
\textbf{Yi Yu}$^{2}$,
\textbf{Ding Wang}$^{2}$,
\textbf{Xi Chen}$^{6}$, \\ 
\textbf{Ben M. Chen}$^{6}$,
\textbf{Yuxuan Liang}$^{5}$,
\textbf{Zhiyong Cui}$^{3}$,
\textbf{Man On Pun}$^{1,*}$, 
\textbf{Yirong Chen}$^{2,}$\thanks{Corresponding authors.} \\ 
$^{1}$ The Chinese University of Hong Kong, Shenzhen \quad
$^{2}$ Shanghai AI Lab \\
$^{3}$ Beihang University \quad
$^{4}$ Nanyang Technological University \\
$^{5}$ The Hong Kong University of Science and Technology (Guangzhou) \\
$^{6}$ The Chinese University of Hong Kong 
}

\date{}

\begin{document}

\maketitle

\begin{abstract}
Traffic scene understanding requires models to reason beyond object recognition, including lane topology, multi-view geometry, temporal evolution, and signal-phase semantics. However, existing traffic-oriented multimodal benchmarks largely emphasize passive visual recognition or isolated video understanding, offering limited support for evaluating structure-aware traffic reasoning under controlled conditions. We introduce \textbf{OmniTraffic}, a controllable generation pipeline and benchmark for spatio-temporal traffic reasoning. Built around 12 real-world intersections reconstructed into editable 3D traffic environments and complemented by surveillance footage from two countries, OmniTraffic supports both controlled and natural-condition evaluation. It defines a three-level task hierarchy spanning scene perception, multi-view and temporal reasoning, and decision support. Using structured traffic metadata, OmniTraffic generates synchronized multi-view VQA samples covering vehicle states, lane functions, view--BEV correspondence, temporal dynamics, and signal-phase analysis, resulting in 8M VQA samples and a 3K human-verified test set. Evaluation of eleven frontier MLLMs reveals a large human--model gap, with the most pronounced failures in topology-grounded and spatio-temporal reasoning tasks. Fine-tuning a lightweight MLLM on simulated OmniTraffic data further improves performance on real-world traffic scenes, demonstrating the value of simulation-generated supervision for traffic-specific multimodal reasoning. Beyond a fixed dataset, OmniTraffic provides an extensible pipeline with configurable intersections, camera views, traffic demands, signal phases, visual conditions, and rare events.
\end{abstract}

\section{Introduction} \label{sec:intro}

Roadside traffic cameras capture rich information about vehicle movements, lane usage, congestion, special vehicles, abnormal events, and signal phases. Multimodal large language models (MLLMs) offer a promising interface for interpreting such data through natural-language queries~\cite{touvron2023llama, achiam2023gpt, yin2024survey, comanici2025gemini, bai2025qwen3, xai2025grok4}, with recent applications in traffic anomaly detection~\cite{orlova2025simplifying, keskar2025evaluating}, congestion analysis~\cite{wu2025trafficinternvl}, traffic question answering~\cite{kuang2025traffic, sheng2025talk2traffic, lu2026emergency}, and signal-aware decision support~\cite{lai2025llmlight, wang2025vlmlight}. However, traffic scene understanding requires models to go beyond generic visual recognition and reason over structured traffic states, including lane topology, cross-view geometry and temporal changes.

Existing traffic benchmarks provide limited controllability for structure-aware evaluation. Real-world videos offer visual diversity, but they provide little control over intersection layouts, signal phases, traffic events, or viewpoint configurations. As a result, evaluations often favor passive perception over reasoning about lane topology, multi-view alignment, and temporal dynamics. Moreover, annotating complex and synchronized traffic states from real footage is labor-intensive and difficult to scale. Controllable simulation offers a practical alternative by directly recording structured metadata, including topology, vehicle states, and signal phases, for traceable VQA generation.

\begin{figure*}
  \includegraphics[width=1\linewidth]{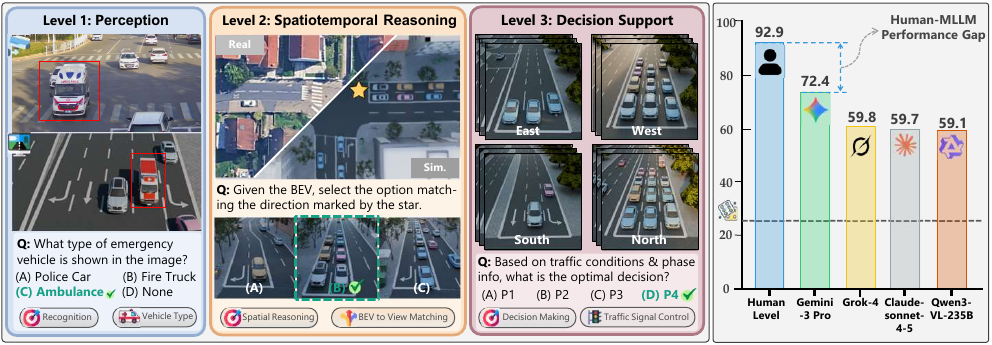}
  \caption{Overview of OmniTraffic. Left: OmniTraffic reconstructs real-world intersections as controllable 3D traffic environments, generates synchronized multi-view observations and structured metadata, and converts them into a three-level VQA hierarchy for perception, spatio-temporal reasoning, and decision support. Right: overall accuracy comparison reveals a significant performance gap between human experts (92.9\%) and the best MLLM (72.4\%).}
  \label{fig_teaser_omnitraffic}
\end{figure*}

In this work, we introduce \textbf{OmniTraffic}, an extensible generation pipeline and benchmark that shifts traffic-oriented VQA from isolated visual recognition toward structured multimodal reasoning. As shown in Fig.~\ref{fig_teaser_omnitraffic}, OmniTraffic organizes traffic understanding into a three-level cognitive hierarchy: \textbf{Perception}, which grounds visual recognition in road structure; \textbf{Spatio-Temporal Reasoning}, which requires cross-view, view--BEV, and temporal reasoning; and \textbf{Decision Support}, which evaluates signal-phase analysis and control-oriented reasoning. To support these tasks, OmniTraffic reconstructs 12 real-world intersections as controllable 3D environments and complements them with real surveillance footage. As illustrated by the Level-2 in Fig.~\ref{fig_teaser_omnitraffic}, real road layouts are reproduced in simulation, enabling controllable scene generation while preserving correspondence to real traffic environments. The pipeline renders synchronized multi-view scenarios, extracts structured traffic metadata, and generates 8M VQA samples together with a 3K human-verified test set. Unlike a closed dataset, OmniTraffic is designed as extensible infrastructure: the released 3D assets, metadata schema, rendering scripts, and VQA generation pipeline allow users to create new samples under configurable camera placements, traffic demands, signal phases, visual conditions, and rare events. OmniTraffic evaluates eleven frontier MLLMs and reveals a substantial human--model gap: the best model reaches 72.4\% overall accuracy, compared with 92.9\% for humans, with pronounced failures on topology-grounded and spatio-temporal reasoning. We further study simulated-to-real transfer by fine-tuning a lightweight MLLM on OmniTraffic dataset and evaluating it on real-world traffic scenes. The resulting overall improvement provides evidence that controllable simulation can offer useful supervision for traffic-specific multimodal reasoning in real-world domain. Our contributions are summarized as follows:

\begin{itemize}[leftmargin=*]
\item We formulate traffic-oriented multimodal understanding as a three-level cognitive hierarchy spanning scene perception, spatio-temporal reasoning, and decision support, with tasks grounded in lane topology and multi-view geometry.
\item We introduce \textbf{OmniTraffic}, a controllable generation pipeline and benchmark built from editable 3D intersections, supporting configurable multi-view rendering, traffic states, and rare events, and producing 8M structured VQA samples with 3K human-verified tests.
\item We evaluate 11 frontier MLLMs, revealing large human--model gaps in structure-aware traffic reasoning, and provide initial evidence that simulated OmniTraffic supervision improves real-world traffic understanding.
\end{itemize}

\section{Related Work}
\begin{table*}[t]
    \centering
    \caption{Comparison of roadside and ego-vehicle traffic benchmarks across data properties and cognitive levels. Perc.: Perception; Rea.: Spatiotemporal Reasoning; Dec.: Decision Support.}
    \label{tab:comparison}
    \resizebox{1\textwidth}{!}{
    \begin{tabular}{l c ccc ccc c}
    \toprule
    \multirow{2}{*}{\textbf{Benchmark}} &
    \multirow{2}{*}{\textbf{Source}} &
    \multicolumn{3}{c}{\textbf{Data Properties}} &
    \multicolumn{3}{c}{\textbf{Cognitive Levels}} &
    \multirow{2}{*}{\textbf{Scale}} \\
    \cmidrule(lr){3-5} \cmidrule(lr){6-8}
    & & \textbf{Topology} & \textbf{Multi-View} & \textbf{Temporal}
        & \textbf{Perc.} & \textbf{Rea.} & \textbf{Dec.} & \\
    \midrule
    \multicolumn{9}{@{}l}{\textbf{\textit{Roadside Benchmarks}}} \\
    \addlinespace[2pt]
    TrafficCAM~\cite{deng2024trafficcam}      & Real     & \xmark & \xmark & \xmark & \cmark & \xmark & \xmark & 58,689    \\
    MTID~\cite{jensen2020presenting}          & Real     & \xmark & \xmark & \cmark & \cmark & \xmark & \xmark & 69,157    \\
    SUTD-TrafficQA~\cite{xu2021sutd}          & Real     & \xmark & \xmark & \cmark & \cmark & \cmark & \xmark & 62,535    \\
    RoadSafe365~\cite{liu2026understanding}   & Real     & \xmark & \xmark & \cmark & \cmark & \cmark & \xmark & 217,000   \\
    AccidentBench~\cite{gu2025accidentbench}  & Real     & \xmark & \xmark & \cmark & \cmark & \cmark & \xmark & 19,000    \\
    TUMTraf VideoQA~\cite{zhou2025tumtraf}    & Real     & \xmark & \xmark & \cmark & \cmark & \cmark & \xmark & 93,000    \\
    MITS~\cite{zhao2025mits}                  & Real     & \xmark & \xmark & \cmark & \cmark & \xmark & \xmark & 5,373,391 \\
    TSBOW~\cite{Huynh2026TSBOW}               & Real     & \xmark & \xmark & \cmark & \cmark & \cmark & \xmark & 3,267,598 \\
    SynTraC~\cite{chen2024syntrac}            & Sim.      & \xmark & \cmark & \cmark & \cmark & \xmark & \cmark & 86,000    \\
    \midrule
    \multicolumn{9}{@{}l}{\textbf{\textit{Ego-vehicle Benchmarks}}} \\
    \addlinespace[2pt]
    nuScenes-QA~\cite{caesar2020nuscenes}     & Real     & \xmark & \xmark & \cmark & \cmark & \cmark & \xmark & 459,941   \\
    DriveLM~\cite{sima2024drivelm}            & Sim./Real & \xmark & \xmark & \cmark & \cmark & \cmark & \cmark & 4,195,000 \\
    \midrule
    \rowcolor{gray!10}
    \textbf{OmniTraffic (Ours)}               & Sim./Real & \cmark & \cmark & \cmark & \cmark & \cmark & \cmark & \textbf{8,092,188} \\
    \bottomrule
    \end{tabular}
    }
\end{table*}

\paragraph{Traffic Benchmarks and Structure-Aware Evaluation.}
Traffic benchmarks can be broadly grouped into ego-vehicle and roadside settings. Ego-vehicle datasets such as BDD100K~\cite{yu2020bdd100k}, nuScenes~\cite{caesar2020nuscenes}, DriveLM~\cite{sima2024drivelm}, and Bench2Drive~\cite{jia2024bench2drive} support perception, prediction, planning, and language-guided driving, while related work on HD mapping~\cite{li2022hdmapnet, liao2023maptr, liao2024maptrv2} and lane topology estimation~\cite{wang2023openlanev2, li2023toponet, li2023lanesegnet} highlights the importance of structured road semantics. Roadside benchmarks instead focus on traffic cameras, ranging from detection under adverse visual conditions~\cite{jensen2020presenting, deng2024trafficcam, xu2024tad, xu2024raod, Huynh2026TSBOW} to video-level event recognition, accident reasoning, and temporal traffic understanding~\cite{xu2021sutd, zhou2025tumtraf, gu2025accidentbench, liu2026understanding, zhao2025mits}. As summarized in Table~\ref{tab:comparison}, existing benchmarks rarely combine lane-level topology, vehicle-to-lane grounding, signal-phase semantics, synchronized multi-view observations, and temporal reasoning within a unified VQA protocol. OmniTraffic addresses this gap by evaluating traffic scenes as structured intersection systems rather than generic images or videos.

\paragraph{MLLMs for Traffic Reasoning and Decision Support.}
MLLMs have recently been applied to traffic scene understanding~\cite{phimsiri2025trafficinternvl}, anomaly detection~\cite{lin2026taur1, zhu2025vau}, lane recognition~\cite{yenikaya2013keeping, zhang2026ladel}, accident causality analysis~\cite{liang2025crashchat, xiu2025traffic}, safety-oriented captioning~\cite{dinh2024trafficvlm, ashqar2025advancing, shi2025scvlm}, driving planning~\cite{tian2024drivevlm, li2025fine, xu2025drivegpt4, zhao2025vlm}, and traffic signal control~\cite{lai2025llmlight, wang2025vlmlight}. However, most adopt task-specific objectives and protocols, leaving model progression from perception to topology-grounded spatio-temporal reasoning and phase-level decision support underexplored. OmniTraffic addresses this gap with a unified traffic cognition hierarchy grounded in lane functions, vehicle--lane assignments, view--BEV correspondence, temporal dynamics, and phase mappings.

\paragraph{Controllable Traffic Generation and Sim-to-Real Transfer.}
Real-world traffic data is difficult to scale for structure-aware evaluation because it requires lane-level topology, vehicle states, camera geometry, temporal correspondences, and phase annotations, while rare events are hard to repeatedly capture. Simulators such as CARLA~\cite{dosovitskiy2017carla}, SUMO~\cite{lopez2018microscopic}, MetaDrive~\cite{li2022metadrive}, SMARTS~\cite{SMARTS}, and LibSignal~\cite{mei2024libsignal}, with benchmarks such as TSLib~\cite{tran2021tslib} and SynTraC~\cite{chen2024syntrac}, support scalable traffic simulation and signal-control research. OmniTraffic extends this paradigm to multimodal traffic understanding through reconstructed intersections, structured metadata, and topology-grounded VQA generation. Fine-tuning results show OmniTraffic supervision improves real-world traffic understanding, extending sim-to-real transfer from policy learning to multimodal reasoning.

\section{OmniTraffic: Benchmark and Generation Pipeline}

\begin{figure*}[!t]
  \centering
  \includegraphics[width=1\linewidth]{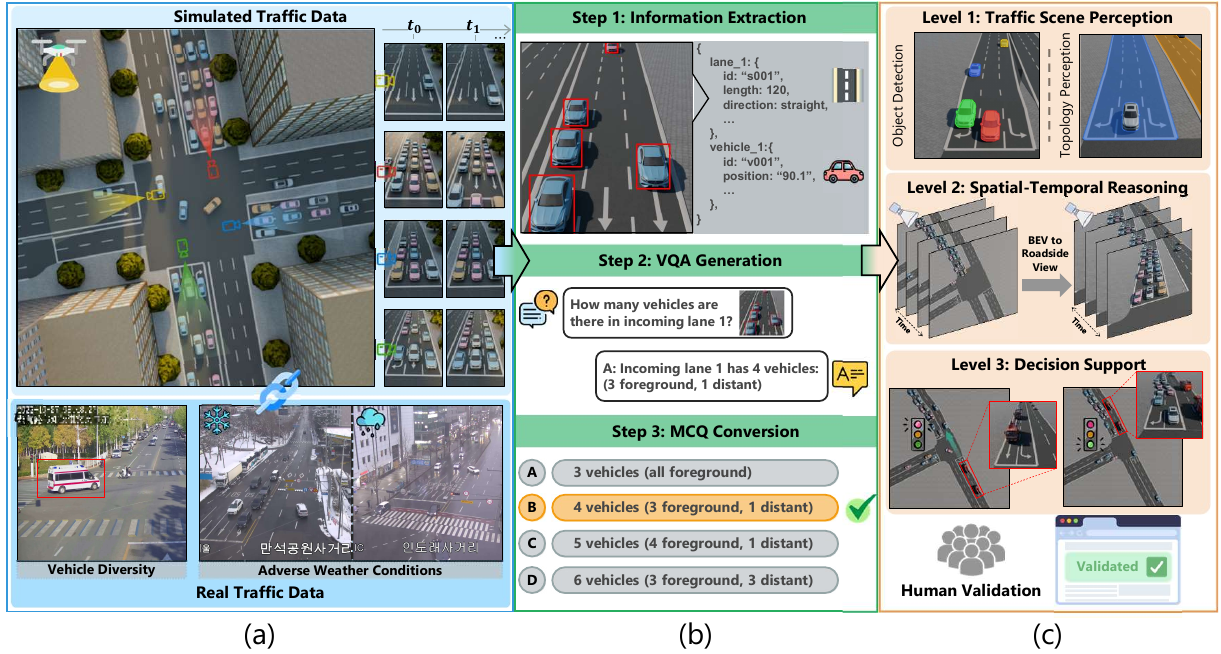}
  \caption{Overview of the OmniTraffic construction pipeline. (a) OmniTraffic reconstructs real-world intersections as controllable 3D traffic environments and complements them with real-world surveillance footage. (b) Simulated images are paired with structured traffic metadata, converted into VQA pairs, and further normalized into multiple-choice questions. (c) The resulting benchmark covers three cognitive levels: perception, spatio-temporal reasoning, and decision support, with human verification for quality control.} 
  \label{fig:pipeline}
\end{figure*}

Fig.~\ref{fig:pipeline} summarizes the construction pipeline of OmniTraffic. Starting from real-world intersection topology, OmniTraffic reconstructs controllable 3D traffic environments, renders synchronized multi-view and temporal observations, records structured traffic-state metadata, and converts these metadata into VQA samples and multiple-choice questions. The resulting benchmark follows a three-level hierarchy covering perception, spatio-temporal reasoning, and decision support. In addition to the simulated data generated from reconstructed intersections, OmniTraffic incorporates real-world surveillance footage as a complementary evaluation domain for assessing model robustness and simulation-to-reality transfer.

\subsection{Controllable Traffic World Construction}

OmniTraffic draws on two complementary data sources, as shown in Fig.~\ref{fig:pipeline}(a). Simulated traffic data enables controllable scene generation, synchronized multi-view observation, temporal continuity, and precise traffic-state metadata for scalable VQA construction. Real-world surveillance data provides an authentic evaluation domain for assessing model robustness, generalization, and sim-to-real transfer under natural visual conditions. Details are provided in Appendix~\ref{app:dataset_construction}.

\begin{figure*}[!t]
  \centering
  \includegraphics[width=0.98\linewidth]{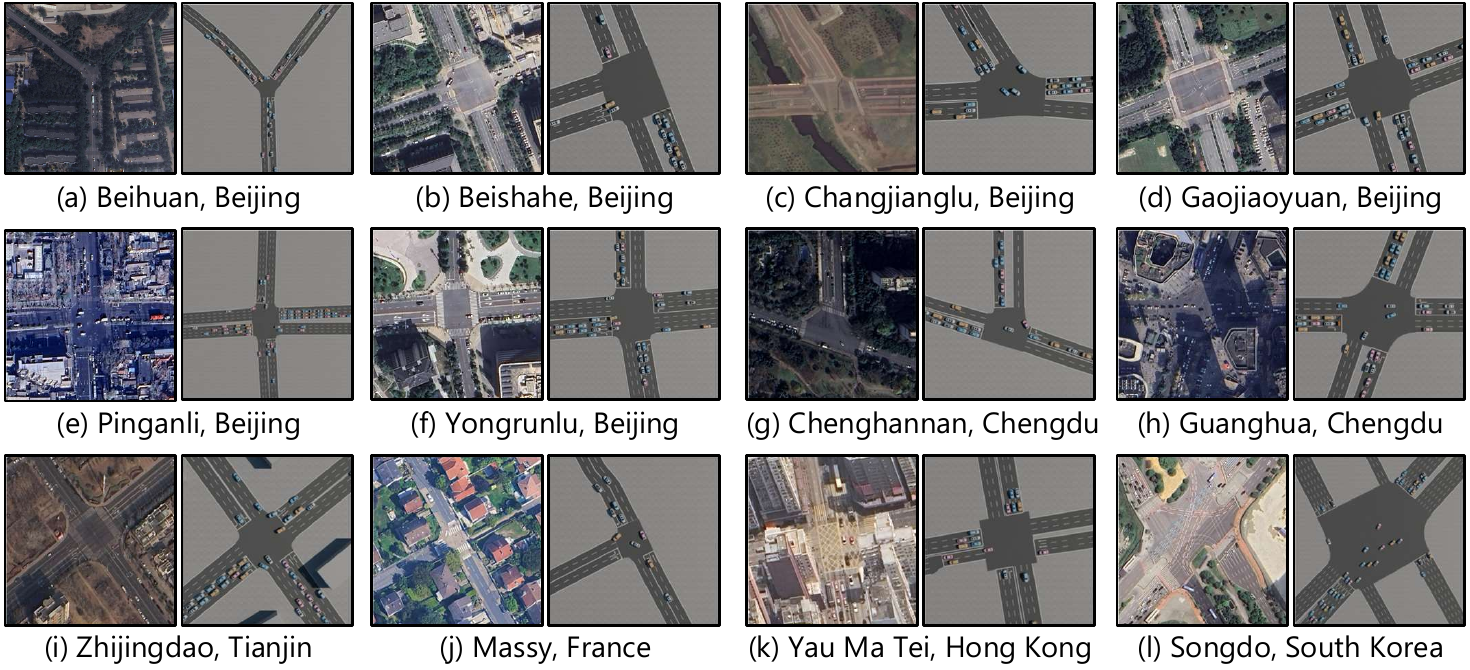}
  \caption{The 12 reconstructed intersections in OmniTraffic. Each pair shows the satellite imagery (left) and the corresponding simulation rendering (right).}
  \label{fig:intersections}
\end{figure*}

\textbf{Simulated Traffic Data.}
We reconstruct 12 real-world intersections as high-fidelity 3D traffic environments. As shown in Fig.~\ref{fig:intersections}, each scene preserves its real-world road topology, including lane geometry, road markings, and signal-phase configurations. Traffic flows are generated through calibrated microsimulation, with controllable special vehicles (\emph{e.g.}, ambulances and police cars), special events (\emph{e.g.}, accidents and road construction), and lighting conditions. A key advantage is \emph{multi-view temporal synchronization}: the same intersection state can be rendered from multiple camera viewpoints at each timestep, producing aligned observations across views and over time. This supports cross-view correspondence, view--BEV mapping, temporal reasoning, and phase-level analysis. It also records structured metadata, including road topology, camera configuration, vehicle states, signal phases, and event annotations, which provides the foundation for the VQA generation pipeline in Sec.~\ref{sec:vqa_generation}. We will release the 3D assets and rendering scripts to support customized data generation with new camera viewpoints, traffic configurations, and environmental conditions.

\begin{figure*}[!t]
  \centering
  \includegraphics[width=0.98\linewidth]{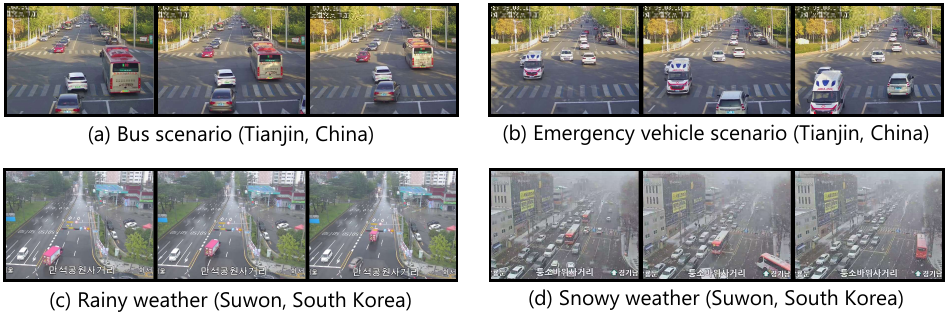}
  \caption{Sample frames from real-world surveillance footage collected in South Korea and Tianjin, China, covering diverse traffic conditions and seasonal variations.}
  \label{fig:realdata}
\end{figure*}

\textbf{Real Traffic Data.}
OmniTraffic also includes real-world surveillance videos from South Korea~\cite{Huynh2026TSBOW} and Tianjin, China, providing an authentic evaluation domain beyond simulation. As shown in Fig.~\ref{fig:realdata}, these videos cover diverse road geometries, traffic patterns, vehicle populations, and seasonal conditions. The real-world portion contains 33.86 hours of video: 32.36 hours from South Korea under sunny, cloudy, rainy, and snowy conditions, and 1.5 hours of continuous 30\,fps recordings from Tianjin.
Since real-world cameras are fixed and do not provide synchronized multi-view coverage of the same traffic state, the real split is used for perception and temporal reasoning tasks that can be reliably constructed from video content. Tasks requiring synchronized multi-view inputs are generated from simulation. This design combines the controllability and exact metadata of simulation with the visual complexity of real roadside footage.

\subsection{Metadata-Driven VQA Generation and Benchmark Curation}
\label{sec:vqa_generation}

After constructing controllable traffic worlds, OmniTraffic converts rendered observations and structured traffic-state metadata into VQA samples. The goal of this stage is to generate questions that are grounded in operational traffic semantics, rather than relying on unconstrained manual question writing. As shown in Fig.~\ref{fig:pipeline}(b), the generation pipeline first produces open-ended question-answer pairs from simulated metadata or real-world annotation, and then converts them into multiple-choice questions for standardized evaluation.

\textbf{Metadata-driven VQA generation.}
For simulated scenes, each rendered frame is paired with structured JSON metadata describing the traffic state. The metadata includes lane length, current signal-phase combination, vehicle-to-lane assignment, each vehicle's longitudinal distance along the lane, vehicle speed, camera viewpoint, and event annotations. These metadata allow OmniTraffic to programmatically generate VQA samples through rule-based templates while preserving exact traceability between the visual observation, the question, and the answer.

The same metadata schema supports different levels of the benchmark hierarchy. For Level 1 perception tasks, vehicle-to-lane assignments and event labels are used to generate questions about vehicle counting, special vehicle recognition, special event recognition, and road infrastructure understanding. For Level 2 spatio-temporal reasoning tasks, synchronized views and temporal state records support questions about cross-view comparison, multi-view localization, view--BEV mapping, and temporal traffic-state changes. For Level 3 decision-support tasks, lane-level traffic states and signal-phase mappings are used to generate phase analysis and phase decision questions. For regular scenes, the phase decision follows a MaxPressure-based rule. For special-event scenes, phases blocked by obstacles are first masked out, and the target phase is selected from the remaining feasible phases. These rules provide interpretable supervision for signal-phase reasoning while keeping the generated answers grounded in scene-specific metadata.

For real-world footage, such structured traffic-state metadata is unavailable. Therefore, three annotators construct and verify VQA samples directly from video content. These real-world questions focus on perception and temporal reasoning tasks that can be reliably annotated from fixed surveillance videos. Since real-world cameras do not provide synchronized multi-view observations of the same traffic state, the real-world split excludes tasks that require synchronized multi-view inputs, such as view--BEV mapping and multi-view phase reasoning.

\textbf{Multiple-choice conversion.}
After obtaining open-ended QA pairs, OmniTraffic normalizes the evaluation format by converting each question into a multiple-choice question. Distractors are sampled from values of the same question type within the same scene, which keeps the options contextually plausible while avoiding trivial choices. For example, vehicle-counting distractors are sampled from other timesteps or camera views of the same scene, temporal prediction distractors are sampled from neighboring frames, and phase-decision options include available signal phases. Duplicate options are removed, and each question is checked to ensure it has a unique correct answer.

\textbf{Quality control and benchmark curation.}
We apply both automatic and manual quality control. Automatic checks verify that numeric answers fall within valid ranges, options are non-duplicated, and each correct answer can be traced back to the corresponding metadata or annotation record. From the full VQA pool, we sample 3,204 items to construct the human-verified benchmark, preserving coverage across task categories and scenes. Three annotators then manually inspect the benchmark to verify image-question alignment, answer correctness, and distractor validity. During this process, questions that are too similar to existing ones are removed and replaced with new samples from the same category and scene group. We additionally develop a web-based evaluation platform where human participants answer the same benchmark questions, enabling a direct comparison between human and model performance across all task levels, see Appendix~\ref{app:benchmark_protocol} for details.

\begin{table}[!t]
\centering
\small
\caption{Overview of the OmniTraffic task hierarchy, including the input format, data source, and evaluation focus of each category.}
\label{tab:task_hierarchy}
\begin{tabularx}{\textwidth}{@{} l l l >{\raggedright\arraybackslash}X @{}}
\toprule
\textbf{Category (Abbr.)} & \textbf{Input} & \textbf{Source} & \textbf{Description} \\ 
\midrule
\rowcolor{l1color!30} 
\multicolumn{4}{@{}l}{\textbf{\textit{L1: Perception}}} \\ \addlinespace[2pt]
Vehicle Counting (Veh.~Count) & Single img & Sim./Real & Count vehicles per image/lane. \\
Special Vehicle Recog.~(Spe.~Veh.) & Single img & Sim./Real & Detect and classify special vehicles. \\
Special Event Recog.~(Spe.~Evt.) & Single img & Sim. & Detect and classify special events. \\
Road Infrastructure (Road Infra.) & Single img & Sim./Real & Identify lane counts and functions. \\
Scene Attribute (Scene Attr.) & Single img & Real & Recognize weather and pedestrians. \\ 
\midrule
\rowcolor{l2color!30} 
\multicolumn{4}{@{}l}{\textbf{\textit{L2: Spatio-Temporal Reasoning}}} \\ \addlinespace[2pt]
Multi-view Comparison (View~Comp.) & Multi-img & Sim. & Compare vehicle counts across views. \\
Multi-view Localization (View~Loc.) & Multi-img & Sim. & Locate special vehicles across views. \\
view-BEV Mapping (view-BEV) & Multi-img & Sim. & Map between perspective and BEV. \\
Temporal Reasoning (Temp.~Reas.) & Multi-img & Sim./Real & Order frames and track queues. \\ 
\midrule
\rowcolor{l3color!30} 
\multicolumn{4}{@{}l}{\textbf{\textit{L3: Decision Support}}} \\ \addlinespace[2pt]
Signal Phase Analysis (Phase~Ana.) & Multi-img & Sim. & Analyze vehicles/events per phase. \\
Signal Phase Decision (Phase~Dec.) & Multi-img & Sim. & Recommend next optimal green phase. \\ 
\bottomrule
\end{tabularx}
\end{table}

\subsection{Task Hierarchy and Benchmark Statistics}
\label{sec:task_statistics}

OmniTraffic contains 8,092,188 VQA pairs in total, generated from 12 reconstructed simulated intersections and real-world surveillance videos from South Korea and Tianjin, China. The real-world portion contains 33.86 hours of video, including 32.36 hours of multi-condition footage from South Korea covering sunny, cloudy, rainy, and snowy scenarios, and 1.5 hours of continuous 30\,fps recordings from Tianjin. The simulated portion provides synchronized multi-view observations, temporal sequences, and structured traffic-state metadata, supporting the full three-level task hierarchy.

Table~\ref{tab:task_hierarchy} summarizes the OmniTraffic hierarchy; details are in Appendix~\ref{app:task_taxonomy}. The benchmark contains eleven task categories organized into three cognitive levels. Level 1 covers single-image perception tasks, including vehicle counting, special vehicle recognition, special event recognition, road infrastructure understanding, and scene attribute recognition. Level 2 covers spatio-temporal reasoning tasks that require multi-image inputs, including multi-view comparison, multi-view localization, view--BEV mapping, and temporal reasoning. Level 3 covers decision-support tasks that require models to analyze signal phases and recommend the next green phase based on multi-view and multi-temporal traffic evidence. Each category contains multiple question variants. For example, vehicle counting ranges from whole-scene counting to lane-specific counting, progressively testing whether MLLMs can connect visual observations with fine-grained traffic topology.

\begin{wrapfigure}{r}{0.6\linewidth}
  \centering
  \includegraphics[width=0.9\linewidth]{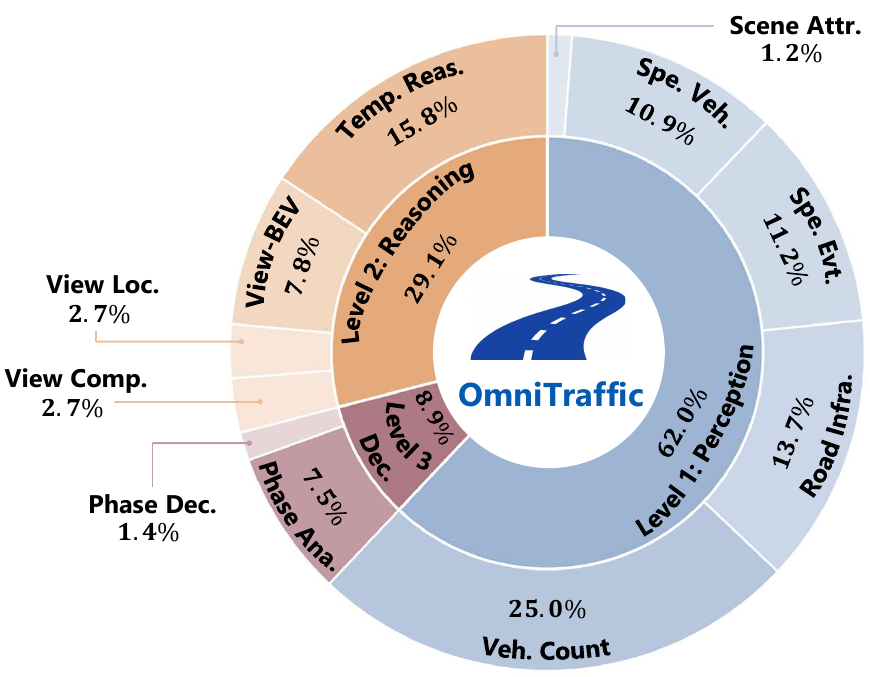}
  \caption{Distribution of the OmniTraffic benchmark across three task levels and eleven categories.}
  \label{fig:distribution}
\end{wrapfigure}

The human-verified benchmark contains 3,204 VQA samples. As shown in Fig.~\ref{fig:distribution}, its distribution follows the three-level hierarchy while preserving coverage across task categories and scenes. Perception tasks form the largest portion because they provide the foundational capabilities for traffic understanding. Spatio-temporal reasoning tasks are fewer but more demanding, as they require models to compare multiple views or timesteps. Decision-support tasks form the most compact subset, reflecting their reliance on higher-order evidence aggregation and signal-phase reasoning. Together, Table~\ref{tab:task_hierarchy} and Fig.~\ref{fig:distribution} show that OmniTraffic is not only large, but organized by increasing cognitive burden: from single-image topology-grounded perception, to multi-image spatio-temporal reasoning, to phase-level decision support. This human-verified benchmark serves as the standardized test set for MLLMs and human participants.

\section{Experiments}

\subsection{Evaluation Protocol}

We evaluate 11 frontier MLLMs, including proprietary models (\emph{e.g.}, GPT-5.2 and Gemini-3-Pro) and open-weight models (\emph{e.g.}, Qwen3-VL-235B), with results reported in Table~\ref{tab:coghier_accuracy}; full model identifiers are provided in Appendix~\ref{app:model_identifiers}. We additionally report a zero-shot human baseline from ten non-expert evaluators. The benchmark aggregates 14 scenarios (12 simulated, 2 real) following the task hierarchy in Table~\ref{tab:task_hierarchy}. For reproducibility, all inferences use greedy decoding ($\tau = 0$, max 500 tokens). Accuracy ($\%$) is measured via exact match against ground-truth labels. To mitigate autoregressive formatting failures, we allow up to three inference attempts per query; invalid responses after three attempts are scored as incorrect.

\subsection{Main Benchmark Results}

As shown in Table~\ref{tab:coghier_accuracy}, Gemini-3-Pro achieves the highest overall model accuracy at 72.4\%, yet remains 20.5 percentage points below the human baseline of 92.9\%. This gap shows that frontier MLLMs still fall substantially short of human performance on traffic-oriented multimodal reasoning. Additional results are provided in Appendix~\ref{app:additional_results}.

However, this gap is non-uniform. Models perform relatively well on tasks with visually salient cues, such as special vehicle recognition and multi-view localization (\emph{e.g.}, best model accuracy is 98.9\% on View Localization vs.\ 93.3\% human). Conversely, performance drops sharply when grounding visual evidence in intersection topology, BEV layouts, temporal traffic evolution, or signal semantics. For instance, accuracy falls to 39.2\% on View--BEV Mapping and 56.4\% on Temporal Reasoning (vs.\ humans' 100.0\% and 94.1\%). These results highlight that structure-aware spatio-temporal reasoning, rather than generic object recognition, is the primary bottleneck.

\begin{table*}[!t]
\centering
\caption{Performance comparison on the OmniTraffic benchmark (\%). $^{\ast}$, $^{\dagger}$, and $^{\ddagger}$ denote the first, second, and third best model results.}
\label{tab:coghier_accuracy}
\resizebox{\textwidth}{!}{
\begin{tabular}{l | *{5}{c} | *{4}{c} | *{2}{c} | c}
\toprule
\multicolumn{1}{c}{\multirow{3}{*}{\textbf{Models}}} &
\multicolumn{5}{c}{\cellcolor{l1color!30}\begin{tabular}{@{}c@{}}L1 Perception\end{tabular}} & 
\multicolumn{4}{c}{\cellcolor{l2color!30}\begin{tabular}{@{}c@{}}L2 Spatio-Temporal \\ Reasoning\end{tabular}} & 
\multicolumn{2}{c}{\cellcolor{l3color!30}\begin{tabular}{@{}c@{}}L3 Decision \\ Support\end{tabular}} & 
\multicolumn{1}{c}{\multirow{3}{*}{\textbf{\begin{tabular}{@{}c@{}}Overall \\ Avg.\end{tabular}}}} \\
& \begin{tabular}{@{}c@{}}Veh. \\ Count\end{tabular} & \begin{tabular}{@{}c@{}}Spe. \\ Veh.\end{tabular} & \begin{tabular}{@{}c@{}}Spe. \\ Evt.\end{tabular} & \begin{tabular}{@{}c@{}}Road \\ Infra.\end{tabular} & \begin{tabular}{@{}c@{}}Scene \\ Attr.\end{tabular}  & \begin{tabular}{@{}c@{}}View \\ Comp.\end{tabular} & \begin{tabular}{@{}c@{}}View \\ Loc.\end{tabular} & \begin{tabular}{@{}c@{}}View- \\ BEV\end{tabular} & \begin{tabular}{@{}c@{}}Temp. \\ Reas.\end{tabular} & \begin{tabular}{@{}c@{}}Phase \\ Ana.\end{tabular} & \begin{tabular}{@{}c@{}}Phase \\ Dec.\end{tabular} \\
\midrule
\multicolumn{13}{@{}l}{\textbf{\textit{Proprietary Models}}} \\
\addlinespace[2pt]
GPT-4o & 41.4 & \third{76.5} & 78.1 & 36.0 & 82.5 & \first{90.9} & 92.0 & \third{32.8} & 48.7 & 74.1 & 56.5 & 55.0 \\
GPT-5.2 & 55.6 & \second{78.5} & \third{78.6} & 41.0 & 80.0 & 76.1 & 89.8 & 25.6 & 43.0 & 67.8 & \third{58.7} & 57.1 \\
Gemini-2.5-Pro & 51.9 & 72.8 & 70.8 & 43.3 & \first{92.5} & 70.5 & \second{96.6} & 9.2 & 22.8 & 71.1 & \second{63.0} & 51.0 \\
Gemini-3-Pro & \first{77.2} & \first{88.2} & \first{90.8} & \first{70.2} & \second{87.5} & \second{89.8} & \first{98.9} & \second{34.0} & 48.9 & \second{81.6} & \first{65.2} & \first{72.4} \\
Claude-Sonnet-4.5 & \second{59.0} & 56.7 & 65.6 & \second{58.1} & 65.0 & \third{87.5} & 86.4 & 29.6 & \second{54.9} & \first{82.0} & \third{58.7} & \third{59.7} \\
Doubao-1.5 & 46.6 & 59.0 & 71.7 & 40.3 & \third{85.0} & 85.2 & 75.0 & 26.0 & \third{50.7} & 72.8 & 56.5 & 53.4 \\
Grok-4 & 47.6 & \second{78.5} & \second{82.5} & 43.1 & 80.0 & 83.0 & 93.2 & \first{39.2} & \first{56.4} & 74.5 & 56.5 & \second{59.8} \\
Qwen-VL-Max & \third{57.5} & 68.5 & 66.7 & \third{55.8} & \third{85.0} & 86.4 & 95.4 & 28.0 & 45.7 & \third{77.8} & \first{65.2} & 59.1 \\
Qwen3-VL-Plus & 50.2 & 68.8 & 70.8 & 38.7 & 77.5 & 63.6 & \third{95.5} & 31.6 & 41.6 & \second{81.6} & \second{63.0} & 54.6 \\
\midrule
\multicolumn{13}{@{}l}{\textbf{\textit{Open-source Models}}} \\
\addlinespace[2pt]
InternVL & 56.0 & 59.6 & 61.9 & 54.4 & 70.0 & 45.5 & 86.4 & 18.0 & 28.7 & 70.3 & 56.5 & 51.4 \\
Qwen3-VL-235B & 55.5 & 67.9 & 73.6 & 48.8 & \second{87.5} & \second{89.8} & 90.9 & 28.4 & 50.5 & 77.4 & \third{58.7} & 59.1 \\
\midrule
\textbf{Human Level} & 93.3 & 96.0 & 100.0 & 86.9 & 93.3 & 100.0 & 93.3 & 100.0 & 94.1 & 88.8 & 55.0 & \textbf{92.9} \\
\bottomrule
\end{tabular}
}
\end{table*}

\begin{figure*}[!ht]
  \centering
  \includegraphics[width=0.99\linewidth]{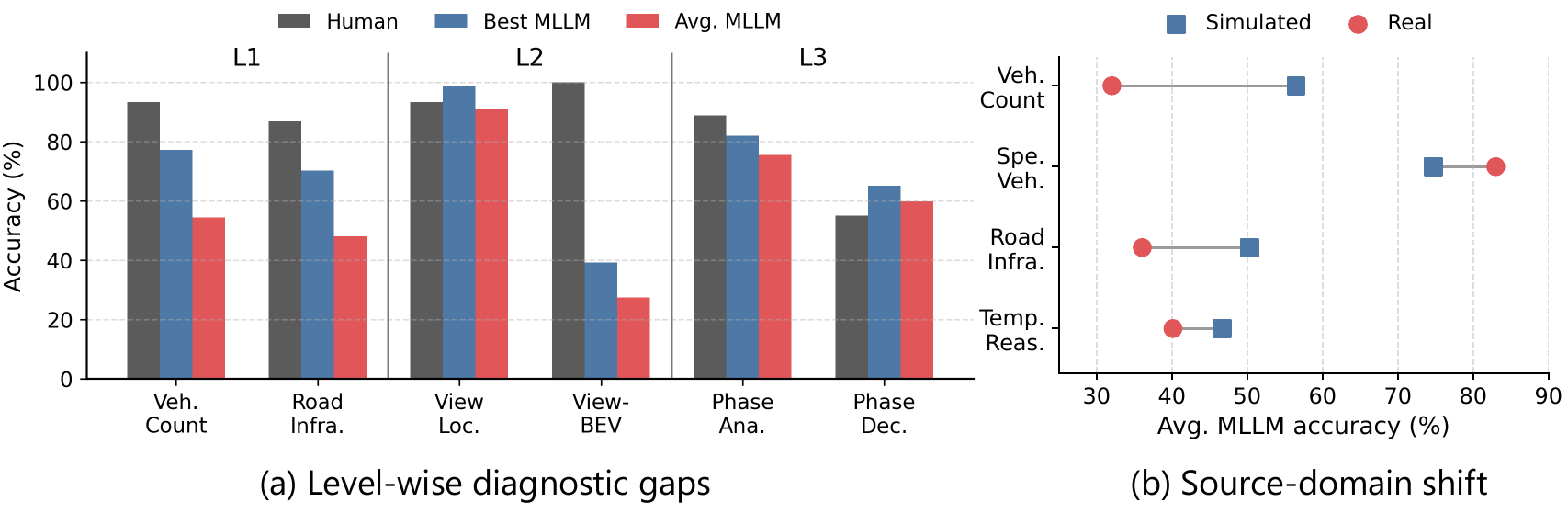}
  \caption{Capability gaps and source-domain shift in OmniTraffic. (a) Accuracy on two representative tasks from each level, aligned with the L1--L3 analysis in the main text. (b) Simulated versus real-world average MLLM accuracy on task categories supported by both sources.}
  \label{fig:capability_gap}
\end{figure*}

Fig.~\ref{fig:capability_gap} further summarizes this capability pattern across simulated and real scenes: the model--human gap is evident in L1 perception tasks requiring lane-level grounding, becomes largest in L2 spatio-temporal reasoning, and appears mixed in L3 decision support. We detail these findings below.

\subsection{Capability Diagnosis}

\textbf{Topology-Grounded Perception Remains Challenging (L1).}
L1 tasks go beyond generic object recognition by requiring models to ground visual evidence in lane topology and road semantics. As shown in Table~\ref{tab:coghier_accuracy} and Fig.~\ref{fig:capability_gap}(a), the strongest model achieves 77.2\% on Vehicle Counting, still far below the human baseline of 93.3\%, indicating difficulty in associating vehicles with specific lanes rather than merely detecting instances. Similarly, for Road Infrastructure understanding, which requires inferring lane functions and semantics from road geometry, Gemini-3-Pro reaches 70.2\%, compared with 86.9\% for humans. These gaps show that general-purpose MLLMs still lack the topology-aware spatial grounding needed for perception-level traffic understanding.

\textbf{Spatio-Temporal Reasoning Exposes the Largest Gaps (L2).}
L2 results reveal a sharp asymmetry: models excel at matching salient visual anchors but fail at structured reasoning. On View Localization, Gemini-3-Pro reaches 98.9\%, and several models surpass the 93.3\% non-expert human baseline, suggesting that MLLMs can track distinctive vehicles across views. However, this ability does not transfer to tasks requiring fine-grained spatial grounding: on View--BEV Mapping, model accuracy drops to 9.2\%--39.2\%, far below humans' 100.0\%, and on Temporal Reasoning, the best model reaches only 56.4\%, compared with 94.1\% for humans. These results identify view--BEV alignment and temporal state reasoning as major bottlenecks. Fig.~\ref{fig:capability_gap}(b) shows a similar pattern under real-world visual conditions: salient Special Vehicle recognition improves, while Vehicle Counting, Road Infrastructure, and Temporal Reasoning decline, indicating that real footage mainly amplifies weaknesses in lane-level topology, queue-state estimation, and temporal reasoning.

\textbf{Decision-Support Tasks Reveal Divergent Human--Model Strengths (L3).}
Unlike L1 and L2, top models perform competitively here. On Signal Phase Analysis, Claude-Sonnet-4.5 and Gemini-3-Pro approach the human baseline (82.0\% and 81.6\% vs.\ 88.8\%). On Signal Phase Decision, Gemini-3-Pro and Qwen-VL-Max (65.2\%) even exceed non-expert humans (55.0\%). However, this reversal should be interpreted cautiously: MLLMs likely exploit traffic-rule knowledge from pretraining, whereas human evaluators lack traffic-engineering backgrounds. MLLMs show promise for phase-level decision support, but their geometric and temporal grounding remains fundamentally limited.

\subsection{Simulation-to-Reality Transfer}

Beyond benchmarking existing models, we examine whether simulated OmniTraffic data improves real-world traffic scene understanding. We use Qwen3-VL-2B-Instruct as a lightweight backbone and fine-tune it with Low-Rank Adaptation (LoRA)~\cite{hu2022lowrank}. The model is trained only on simulated OmniTraffic samples and evaluated on real-world traffic VQA, testing whether controllable simulation provides transferable supervision. Detailed settings are provided in Appendix~\ref{app:sim2real_detail}.

\begin{wraptable}{r}{0.50\linewidth}
  \centering
  \caption{Simulated-to-real transfer results on real-world traffic scene understanding tasks (\%).}
  \label{tab:results_single_column}
  \resizebox{0.85\linewidth}{!}{%
  \begin{tabular}{lcc|c}
    \toprule
    \textbf{Model} & \textbf{Korea} & \textbf{Tianjin} & \textbf{All} \\
    \midrule
    \multicolumn{4}{@{}l}{\textbf{\textit{Proprietary Models}}} \\
    \addlinespace[2pt]
    Grok-4 & 54.9 & 53.7 & 54.1 \\
    GPT-5.2 & 48.4 & 48.9 & 48.7 \\
    Gemini-3-Pro & 60.7 & 54.5 & 56.6 \\
    Claude-4.5 & 50.8 & 52.4 & 51.8 \\
    Doubao-1.5 & 54.1 & 47.2 & 49.6 \\
    Qwen-VL-Max & 51.6 & 49.8 & 50.4 \\
    \midrule
    \multicolumn{4}{@{}l}{\textbf{\textit{Open-source Models}}} \\
    \addlinespace[2pt]
    Qwen3-235B & 53.3 & 52.8 & 53.0 \\
    Qwen3-8B & 49.2 & 50.2 & 49.9 \\
    Qwen3-2B & 39.3 & 51.9 & 47.6 \\
    \midrule
    \multicolumn{4}{@{}l}{\textbf{\textit{OmniTraffic Fine-tuning}}} \\
    \addlinespace[2pt]
    Qwen3-2B+FT & \textbf{52.5} & \textbf{54.5} & \textbf{53.8} \\
    \bottomrule
  \end{tabular}
    }
  \vspace{-1.2em}
\end{wraptable}

As shown in Table~\ref{tab:results_single_column}, the fine-tuned 2B model achieves 53.8\% overall accuracy on real-world scenes, improving its base counterpart by +6.2 points (47.6\%). The gain is scene-dependent, with a larger improvement on Korea (+13.2 points) than on Tianjin (+2.6 points). Notably, the simulation-tuned 2B model surpasses Qwen3-8B (49.9\%) and slightly exceeds Qwen3-235B (53.0\%), suggesting that OmniTraffic provides effective synthetic supervision without real-world training annotations.

This overall improvement indicates that simulation can teach domain-invariant traffic structures, including lane topology, queue evolution, and rule-like semantic associations. However, simulation-only fine-tuning remains limited for appearance-sensitive real-world understanding, such as viewpoint variation, occlusion, illumination, and local visual details; category-level evidence is provided in Appendix~\ref{app:transfer_results}. Together, these results show that OmniTraffic is not only a benchmark, but also a controlled generation pipeline for training and validating sim-to-real traffic-scene understanding models.

\section{Conclusion}

We presented \textbf{OmniTraffic}, a controllable generation pipeline and benchmark for spatio-temporal traffic reasoning. Built from reconstructed real-world intersections and complemented by real-world surveillance footage, OmniTraffic generates metadata-grounded VQA samples across perception, spatio-temporal reasoning, and decision support. Experiments with eleven frontier MLLMs reveal a substantial human--model gap, particularly on topology-grounded and temporal reasoning tasks such as view--BEV mapping and traffic-state evolution. Fine-tuning a lightweight MLLM on simulated OmniTraffic data improves overall real-world performance, providing evidence that controllable simulation can offer useful supervision for traffic-specific multimodal reasoning. OmniTraffic provides an extensible benchmark and generation pipeline for future research on structure-aware traffic intelligence.

\bibliographystyle{unsrt}
\bibliography{reference}

@techreport{xai2025grok4, 
  title={Grok 4 Model Card}, 
  author={{xAI}}, 
  year={2025}, 
  month={August}, 
  institution={xAI}, 
  url={https://data.x.ai/2025-08-20-grok-4-model-card.pdf} 
}

@inproceedings{lopez2018microscopic,
  title={Microscopic traffic simulation using sumo},
  author={Lopez, Pablo Alvarez and Behrisch, Michael and Bieker-Walz, Laura and Erdmann, Jakob and Fl{\"o}tter{\"o}d, Yun-Pang and Hilbrich, Robert and L{\"u}cken, Leonhard and Rummel, Johannes and Wagner, Peter and Wie{\ss}ner, Evamarie},
  booktitle={2018 21st international conference on intelligent transportation systems (ITSC)},
  pages={2575--2582},
  year={2018},
  organization={Ieee}
}

@article{li2022metadrive,
  title={Metadrive: Composing diverse driving scenarios for generalizable reinforcement learning},
  author={Li, Quanyi and Peng, Zhenghao and Feng, Lan and Zhang, Qihang and Xue, Zhenghai and Zhou, Bolei},
  journal={IEEE transactions on pattern analysis and machine intelligence},
  volume={45},
  number={3},
  pages={3461--3475},
  year={2022},
  publisher={IEEE}
}

@article{mei2024libsignal,
  title={Libsignal: an open library for traffic signal control},
  author={Mei, Hao and Lei, Xiaoliang and Da, Longchao and Shi, Bin and Wei, Hua},
  journal={Machine Learning},
  volume={113},
  number={8},
  pages={5235--5271},
  year={2024},
  publisher={Springer}
}

@inproceedings{phimsiri2025trafficinternvl,
  title={TrafficInternVL: Spatially-Guided Fine-Tuning with Caption Refinement for Fine-Grained Traffic Safety Captioning and Visual Question Answering},
  author={Phimsiri, Sasin and Sunpawatr, Sarut and Cherdchusakulchai, Riu and Kiawjak, Pornprom and Tosawadi, Teepakorn and Tungjitnob, Suchat and Trairattanapa, Visarut and Vatathanavaro, Supawit and Kudisthalert, Wasu and Utintu, Chaitat and others},
  booktitle={Proceedings of the IEEE/CVF International Conference on Computer Vision},
  pages={5299--5306},
  year={2025}
}

@misc{SMARTS,
    title={SMARTS: Scalable Multi-Agent Reinforcement Learning Training School for Autonomous Driving},
    author={Ming Zhou and Jun Luo and Julian Villella and Yaodong Yang and David Rusu and Jiayu Miao and Weinan Zhang and Montgomery Alban and Iman Fadakar and Zheng Chen and Aurora Chongxi Huang and Ying Wen and Kimia Hassanzadeh and Daniel Graves and Dong Chen and Zhengbang Zhu and Nhat Nguyen and Mohamed Elsayed and Kun Shao and Sanjeevan Ahilan and Baokuan Zhang and Jiannan Wu and Zhengang Fu and Kasra Rezaee and Peyman Yadmellat and Mohsen Rohani and Nicolas Perez Nieves and Yihan Ni and Seyedershad Banijamali and Alexander Cowen Rivers and Zheng Tian and Daniel Palenicek and Haitham bou Ammar and Hongbo Zhang and Wulong Liu and Jianye Hao and Jun Wang},
    url={https://arxiv.org/abs/2010.09776},
    primaryClass={cs.MA},
    booktitle={Proceedings of the 4th Conference on Robot Learning (CoRL)},
    year={2020},
    month={11}
}

@inproceedings{tran2021tslib,
  title={{TSLib}: A unified traffic signal control framework using deep reinforcement learning and benchmarking},
  author={Tran, Toan V and Doan, Thanh-Nam and Sartipi, Mina},
  booktitle={2021 IEEE international conference on big data (Big Data)},
  pages={1739--1747},
  year={2021},
  organization={IEEE}
}

@inproceedings{chen2024syntrac,
  title={SynTrac: a synthetic dataset for traffic signal control from traffic monitoring cameras},
  author={Chen, Tiejin and Shirke, Prithvi and Chakravarthi, Bharatesh and Vaghela, Arpitsinh and Da, Longchao and Lu, Duo and Yang, Yezhou and Wei, Hua},
  booktitle={2024 IEEE 27th International Conference on Intelligent Transportation Systems (ITSC)},
  pages={2386--2391},
  year={2024},
  organization={IEEE}
}

@inproceedings{dosovitskiy2017carla,
  title={CARLA: An open urban driving simulator},
  author={Dosovitskiy, Alexey and Ros, German and Codevilla, Felipe and Lopez, Antonio and Koltun, Vladlen},
  booktitle={Conference on robot learning},
  pages={1--16},
  year={2017},
  organization={PMLR}
}

@inproceedings{yu2020bdd100k,
  title={Bdd100k: A diverse driving dataset for heterogeneous multitask learning},
  author={Yu, Fisher and Chen, Haofeng and Wang, Xin and Xian, Wenqi and Chen, Yingying and Liu, Fangchen and Madhavan, Vashisht and Darrell, Trevor},
  booktitle={Proceedings of the IEEE/CVF conference on computer vision and pattern recognition},
  pages={2636--2645},
  year={2020}
}

@inproceedings{caesar2020nuscenes,
  title={{nuScenes}: A multimodal dataset for autonomous driving},
  author={Caesar, Holger and Bankiti, Varun and Lang, Alex H and Vora, Sourabh and Liong, Venice Erin and Xu, Qiang and Krishnan, Anush and Pan, Yu and Baldan, Giancarlo and Beijbom, Oscar},
  booktitle={Proceedings of the IEEE/CVF conference on computer vision and pattern recognition},
  pages={11621--11631},
  year={2020}
}

@inproceedings{sima2024drivelm,
  title={{DriveLM}: Driving with graph visual question answering},
  author={Sima, Chonghao and Renz, Katrin and Chitta, Kashyap and Chen, Li and Zhang, Hanxue and Xie, Chengen and Bei{\ss}wenger, Jens and Luo, Ping and Geiger, Andreas and Li, Hongyang},
  booktitle={European conference on computer vision},
  pages={256--274},
  year={2024},
  organization={Springer}
}

@inproceedings{xu2025drivegpt4,
  title={Drivegpt4-v2: Harnessing large language model capabilities for enhanced closed-loop autonomous driving},
  author={Xu, Zhenhua and Bai, Yan and Zhang, Yujia and Li, Zhuoling and Xia, Fei and Wong, Kwan-Yee K and Wang, Jianqiang and Zhao, Hengshuang},
  booktitle={Proceedings of the Computer Vision and Pattern Recognition Conference},
  pages={17261--17270},
  year={2025}
}

@article{ashqar2025advancing,
  title={Advancing object detection in transportation with multimodal large language models (MLLMs): A comprehensive review and empirical testing},
  author={Ashqar, Huthaifa I and Jaber, Ahmed and Alhadidi, Taqwa I and Elhenawy, Mohammed},
  journal={Computation},
  volume={13},
  number={6},
  pages={133},
  year={2025},
  publisher={MDPI}
}

@inproceedings{liao2023maptr,
  title={{MapTR}: Structured Modeling and Learning for Online Vectorized HD Map Construction},
  author={Liao, Bencheng and Chen, Shaoyu and Wang, Xinggang and Cheng, Tianheng and Zhang, Qian and Liu, Wenyu and Huang, Chang},
  booktitle={International Conference on Learning Representations},
  year={2023}
}

@article{liao2024maptrv2,
  title={Maptrv2: An end-to-end framework for online vectorized hd map construction},
  author={Liao, Bencheng and Chen, Shaoyu and Zhang, Yunchi and Jiang, Bo and Zhang, Qian and Liu, Wenyu and Huang, Chang and Wang, Xinggang},
  journal={International Journal of Computer Vision},
  pages={1--23},
  year={2024},
  publisher={Springer}
}

@inproceedings{wang2023openlanev2,
  title={{OpenLane-V2}: A Topology Reasoning Benchmark for Unified 3D HD Mapping}, 
  author={Wang, Huijie and Li, Tianyu and Li, Yang and Chen, Li and Sima, Chonghao and Liu, Zhenbo and Wang, Bangjun and Jia, Peijin and Wang, Yuting and Jiang, Shengyin and Wen, Feng and Xu, Hang and Luo, Ping and Yan, Junchi and Zhang, Wei and Li, Hongyang},
  booktitle={NeurIPS},
  year={2023}
}

@article{li2023toponet,
  title={Graph-based Topology Reasoning for Driving Scenes},
  author={Li, Tianyu and Chen, Li and Wang, Huijie and Li, Yang and Yang, Jiazhi and Geng, Xiangwei and Jiang, Shengyin and Wang, Yuting and Xu, Hang and Xu, Chunjing and Yan, Junchi and Luo, Ping and Li, Hongyang},
  journal={arXiv preprint arXiv:2304.05277},
  year={2023}
}

@inproceedings{li2023lanesegnet,
  title={{LaneSegNet}: Map Learning with Lane Segment Perception for Autonomous Driving},
  author={Li, Tianyu and Jia, Peijin and Wang, Bangjun and Chen, Li and Jiang, Kun and Yan, Junchi and Li, Hongyang},
  booktitle={ICLR},
  year={2024}
}

@article{jia2024bench2drive,
  title={Bench2drive: Towards multi-ability benchmarking of closed-loop end-to-end autonomous driving},
  author={Jia, Xiaosong and Yang, Zhenjie and Li, Qifeng and Zhang, Zhiyuan and Yan, Junchi},
  journal={Advances in Neural Information Processing Systems},
  volume={37},
  pages={819--844},
  year={2024}
}

@inproceedings{li2022hdmapnet,
  title={Hdmapnet: An online hd map construction and evaluation framework},
  author={Li, Qi and Wang, Yue and Wang, Yilun and Zhao, Hang},
  booktitle={2022 International Conference on Robotics and Automation (ICRA)},
  pages={4628--4634},
  year={2022},
  organization={IEEE}
}

@article{Huynh2026TSBOW, 
    title={TSBOW: Traffic Surveillance Benchmark for Occluded Vehicles Under Various Weather Conditions}, 
    volume={40}, 
    url={https://ojs.aaai.org/index.php/AAAI/article/view/37439}, 
    DOI={10.1609/aaai.v40i7.37439}, 
    number={7}, 
    journal={Proceedings of the AAAI Conference on Artificial Intelligence}, 
    author={Huynh, Ngoc Doan-Minh and Tran, Duong Nguyen-Ngoc and Pham, Long Hoang and Tran, Tai Huu-Phuong and Jeon, Hyung-Joon and Nguyen, Huy-Hung and Khac Vu, Duong and Jeon, Hyung-Min and Phan, Son Hong and Pham-Nam Ho, Quoc and Tran, Chi Dai and Khanh, Trinh Le Ba and Jeon, Jae Wook}, 
    year={2026}, 
    month={Mar.}, 
    pages={5239-5247} 
}

@article{xu2024tad,
  title={TAD: A large-scale benchmark for traffic accidents detection from video surveillance},
  author={Xu, Yajun and Hu, Huan and Huang, Chuwen and Nan, Yibing and Liu, Yuyao and Wang, Kai and Liu, Zhaoxiang and Lian, Shiguo},
  journal={IEEE Access},
  volume={13},
  pages={2018--2033},
  year={2024},
  publisher={IEEE}
}

@article{xu2024raod,
  title={Raod: A benchmark for road abandoned object detection from video surveillance},
  author={Xu, Yajun and Hu, Huan and Zhu, Xiaoya and Nan, Yibing and Wang, Kai and Liu, Zhaoxiang and Lian, Shiguo},
  journal={IEEE Access},
  volume={12},
  pages={123985--123994},
  year={2024},
  publisher={IEEE}
}

@article{deng2024trafficcam,
  title={{TrafficCAM}: A versatile dataset for traffic flow segmentation},
  author={Deng, Zhongying and Cheng, Yanqi and Liu, Lihao and Wang, Shujun and Ke, Rihuan and Sch{\"o}nlieb, Carola-Bibiane and Aviles-Rivero, Angelica I},
  journal={IEEE Transactions on Intelligent Transportation Systems},
  volume={26},
  number={2},
  pages={2747--2759},
  year={2024},
  publisher={IEEE}
}

@inproceedings{jensen2020presenting,
  title={Presenting the multi-view traffic intersection dataset (MTID): A detailed traffic-surveillance dataset},
  author={Jensen, Morten B and M{\o}gelmose, Andreas and Moeslund, Thomas B},
  booktitle={2020 IEEE 23rd International Conference on Intelligent Transportation Systems (ITSC)},
  pages={1--6},
  year={2020},
  organization={IEEE}
}

@inproceedings{xu2021sutd,
  title={Sutd-trafficqa: A question answering benchmark and an efficient network for video reasoning over traffic events},
  author={Xu, Li and Huang, He and Liu, Jun},
  booktitle={Proceedings of the IEEE/CVF conference on computer vision and pattern recognition},
  pages={9878--9888},
  year={2021}
}

@inproceedings{zhou2025tumtraf,
  title={TUMTraf VideoQA: Dataset and Benchmark for Unified Spatio-Temporal Video Understanding in Traffic Scenes},
  author={Zhou, Xingcheng and Larintzakis, Konstantinos and Guo, Hao and Zimmer, Walter and Liu, Mingyu and Cao, Hu and Zhang, Jiajie and Lakshminarasimhan, Venkatnarayanan and Strand, Leah and Knoll, Alois},
  booktitle={Forty-second International Conference on Machine Learning},
  year={2025}
}

@article{zhao2025mits,
  title={{MITS}: A large-scale multimodal benchmark dataset for Intelligent Traffic Surveillance},
  author={Zhao, Kaikai and Liu, Zhaoxiang and Wang, Peng and Wang, Xin and Ma, Zhicheng and Xu, Yajun and Zhang, Wenjing and Nan, Yibing and Wang, Kai and Lian, Shiguo},
  journal={Image and Vision Computing},
  pages={105736},
  year={2025},
  publisher={Elsevier}
}

@article{liu2026understanding,
  title={Understanding Real-World Traffic Safety through RoadSafe365 Benchmark},
  author={Liu, Xinyu and Jacob, Darryl C and Liu, Yuxin and Du, Xinsong and Ye, Muchao and Zhou, Bolei and He, Pan},
  journal={arXiv preprint arXiv:2602.07212},
  year={2026}
}

@article{gu2025accidentbench,
  title={AccidentBench: Benchmarking Multimodal Understanding and Reasoning in Vehicle Accidents and Beyond},
  author={Gu, Shangding and Wang, Xiaohan and Ying, Donghao and Zhao, Haoyu and Yang, Runing and Jin, Ming and Li, Boyi and Pavone, Marco and Yeung-Levy, Serena and Wang, Jun and others},
  journal={arXiv preprint arXiv:2509.26636},
  year={2025}
}

@article{lin2026taur1,
  title={TAU-R1: Visual Language Model for Traffic Anomaly Understanding},
  author={Lin, Yuqiang and Chen, Kehua and Lockyer, Sam and Yadav, Arjun and Sui, Mingxuan and Zhang, Shucheng and Shi, Yan and Wang, Bingzhang and Zhang, Yuang and Zarbock, Markus and Stanek, Florain and Evans, Adrian and Li, Wenbin and Wang, Yinhai and Zhang, Nic},
  journal={arXiv preprint arXiv:2603.19098},
  year={2026}
}

@inproceedings{dinh2024trafficvlm,
  title={{TrafficVLM}: A controllable visual language model for traffic video captioning},
  author={Dinh, Quang Minh and Ho, Minh Khoi and Dang, Anh Quan and Tran, Hung Phong},
  booktitle={Proceedings of the IEEE/CVF conference on computer vision and pattern recognition},
  pages={7134--7143},
  year={2024}
}

@article{zhao2025vlm,
  title={{VLM-Driver}: Human-Like Autonomous Driving Decision-Making via Vision Language Model},
  author={Zhao, Rui and Yuan, Qirui and Li, Jinyu and Wang, Zhiqiang and Li, Yun and Gao, Zhenhai and Hu, Hongyu and Gao, Fei},
  journal={IEEE Transactions on Vehicular Technology},
  year={2025},
  publisher={IEEE}
}

@article{tian2024drivevlm,
  title={{DriveVLM}: The convergence of autonomous driving and large vision-language models},
  author={Tian, Xiaoyu and Gu, Junru and Li, Bailin and Liu, Yicheng and Wang, Yang and Zhao, Zhiyong and Zhan, Kun and Jia, Peng and Lang, Xianpeng and Zhao, Hang},
  journal={arXiv preprint arXiv:2402.12289},
  year={2024}
}

@article{liang2025crashchat,
  title={{CrashChat}: A Multimodal Large Language Model for Multitask Traffic Crash Video Analysis},
  author={Liang, Kaidi and Li, Ke and Hu, Xianbiao and Qin, Ruwen},
  journal={arXiv preprint arXiv:2512.18878},
  year={2025}
}

@article{xiu2025traffic,
  title={Traffic-MLLM: A Spatio-Temporal MLLM with Retrieval-Augmented Generation for Causal Inference in Traffic},
  author={Xiu, Waikit and Lu, Qiang and Li, Xiying and Hu, Chen and Sun, Shengbo},
  journal={arXiv preprint arXiv:2509.11165},
  year={2025}
}

@article{zhang2026ladel,
  title={LaDeL: Lane detection via multimodal large language model with visual instruction tuning},
  author={Zhang, Yun and Cheng, Xin and Zhou, Zhou and Zhou, Jingmei and Yang, Tong},
  journal={Journal of Visual Communication and Image Representation},
  pages={104704},
  year={2026},
  publisher={Elsevier}
}

@article{zhu2025vau,
  title={{VAU-R1}: Advancing video anomaly understanding via reinforcement fine-tuning},
  author={Zhu, Liyun and Chen, Qixiang and Shen, Xi and Cun, Xiaodong},
  journal={arXiv preprint arXiv:2505.23504},
  year={2025}
}

@inproceedings{wang2025vlmlight,
  title={VLMLight: Safety-Critical Traffic Signal Control via Vision-Language Meta-Control and Dual-Branch Reasoning Architecture},
  author={Wang, Maonan and Chen, Yirong and Pang, Aoyu and Cai, Yuxin and Chen, Chung Shue and Kan, Yuheng and Pun, Man-On}, 
  booktitle={Advances in Neural Information Processing Systems (NeurIPS)},
  year={2025},
  url={https://neurips.cc/virtual/2025/loc/san-diego/poster/120348}
}

@article{yin2024survey,
  title={A survey on multimodal large language models},
  author={Yin, Shukang and Fu, Chaoyou and Zhao, Sirui and Li, Ke and Sun, Xing and Xu, Tong and Chen, Enhong},
  journal={National Science Review},
  volume={11},
  number={12},
  pages={nwae403},
  year={2024},
  publisher={Oxford University Press}
}

@article{kuang2025traffic,
  title={Traffic-IT: Enhancing traffic scene understanding for multimodal large language models},
  author={Kuang, Senyun and Liu, Yang and Qu, Xiaobo and Wei, Yintao},
  journal={Transportation Research Part C: Emerging Technologies},
  volume={180},
  pages={105325},
  year={2025},
  publisher={Elsevier}
}

@inproceedings{shi2025scvlm,
  title={{ScVLM}: Enhancing vision-language model for safety-critical event understanding},
  author={Shi, Liang and Jiang, Boyu and Zeng, Tong and Guo, Feng},
  booktitle={Proceedings of the Winter Conference on Applications of Computer Vision},
  pages={1061--1071},
  year={2025}
}

@article{yenikaya2013keeping,
  title={Keeping the vehicle on the road: A survey on on-road lane detection systems},
  author={Yenikaya, Sibel and Yenikaya, G{\"o}khan and D{\"u}ven, Ekrem},
  journal={ACM Computing Surveys (Csur)},
  volume={46},
  number={1},
  pages={1--43},
  year={2013},
  publisher={ACM New York, NY, USA}
}

@inproceedings{hu2022lowrank,
  author = {Hu, Edward J. and Shen, Yelong and Wallis, Phillip and Allen-Zhu, Zeyuan and Li, Yuanzhi and Wang, Shean and Wang, Lu and Chen, Weizhu},
  booktitle = {ICLR},
  title = {LoRA: Low-Rank Adaptation of Large Language Models.},
  url = {http://dblp.uni-trier.de/db/conf/iclr/iclr2022.html#HuSWALWWC22},
  year = {2022}
}

@inproceedings{sheng2025talk2traffic,
  title={Talk2traffic: Interactive and editable traffic scenario generation for autonomous driving with multimodal large language model},
  author={Sheng, Zihao and Huang, Zilin and Qu, Yansong and Leng, Yue and Chen, Sikai},
  booktitle={Proceedings of the IEEE/CVF Conference on Computer Vision and Pattern Recognition},
  pages={3797--3806},
  year={2025}
}

@article{lu2026emergency,
  title={Emergency Events Traffic Flow Forecasting Using Text-Prompt-Guided Multimodal Large Language Models},
  author={Lu, Yaxuan and Huo, Guangyu and Cui, Xiaohui and Wang, Boyue and Zhang, Yong and Cui, Zhiyong},
  journal={IEEE Transactions on Intelligent Transportation Systems},
  year={2026},
  publisher={IEEE}
}

@article{bai2025qwen3,
  title={Qwen3-vl technical report},
  author={Bai, Shuai and Cai, Yuxuan and Chen, Ruizhe and Chen, Keqin and Chen, Xionghui and Cheng, Zesen and Deng, Lianghao and Ding, Wei and Gao, Chang and Ge, Chunjiang and others},
  journal={arXiv preprint arXiv:2511.21631},
  year={2025}
}

@article{touvron2023llama,
  title={Llama: Open and efficient foundation language models},
  author={Touvron, Hugo and Lavril, Thibaut and Izacard, Gautier and Martinet, Xavier and Lachaux, Marie-Anne and Lacroix, Timoth{\'e}e and Rozi{\`e}re, Baptiste and Goyal, Naman and Hambro, Eric and Azhar, Faisal and others},
  journal={arXiv preprint arXiv:2302.13971},
  year={2023}
}

@article{achiam2023gpt,
  title={Gpt-4 technical report},
  author={Achiam, Josh and Adler, Steven and Agarwal, Sandhini and Ahmad, Lama and Akkaya, Ilge and Aleman, Florencia Leoni and Almeida, Diogo and Altenschmidt, Janko and Altman, Sam and Anadkat, Shyamal and others},
  journal={arXiv preprint arXiv:2303.08774},
  year={2023}
}

@article{comanici2025gemini,
  title={Gemini 2.5: Pushing the frontier with advanced reasoning, multimodality, long context, and next generation agentic capabilities},
  author={Comanici, Gheorghe and Bieber, Eric and Schaekermann, Mike and Pasupat, Ice and Sachdeva, Noveen and Dhillon, Inderjit and Blistein, Marcel and Ram, Ori and Zhang, Dan and Rosen, Evan and others},
  journal={arXiv preprint arXiv:2507.06261},
  year={2025}
}

@inproceedings{orlova2025simplifying,
  title={Simplifying traffic anomaly detection with video foundation models},
  author={Orlova, Svetlana and Kerssies, Tommie and Englert, Brun{\'o} B and Dubbelman, Gijs},
  booktitle={Proceedings of the IEEE/CVF International Conference on Computer Vision},
  pages={852--862},
  year={2025}
}

@inproceedings{lai2025llmlight,
  title={LLMLight: Large language models as traffic signal control agents},
  author={Lai, Siqi and Xu, Zhao and Zhang, Weijia and Liu, Hao and Xiong, Hui},
  booktitle={Proceedings of the 31st ACM SIGKDD Conference on Knowledge Discovery and Data Mining V. 1},
  pages={2335--2346},
  year={2025}
}

@inproceedings{wu2025trafficinternvl,
  title={TrafficInternVL: Understanding Traffic Scenarios with Vision-Language Models},
  author={Wu, Hsiu-Fu and Yang, Ya-Ting and Chen, Yung-Ter and Chou, I-Fan},
  booktitle={Proceedings of the IEEE/CVF International Conference on Computer Vision},
  pages={5229--5236},
  year={2025}
}

@inproceedings{keskar2025evaluating,
  title={Evaluating multimodal vision-language model prompting strategies for visual question answering in road scene understanding},
  author={Keskar, Aryan and Perisetla, Srinivasa and Greer, Ross},
  booktitle={Proceedings of the Winter Conference on Applications of Computer Vision},
  pages={1027--1036},
  year={2025}
}

@inproceedings{li2025fine,
  title={Fine-grained evaluation of large vision-language models in autonomous driving},
  author={Li, Yue and Tian, Meng and Lin, Zhenyu and Zhu, Jiangtong and Zhu, Dechang and Liu, Haiqiang and Zhang, Yueyi and Xiong, Zhiwei and Zhao, Xinhai},
  booktitle={Proceedings of the IEEE/CVF International Conference on Computer Vision},
  pages={9431--9442},
  year={2025}
}

\clearpage
\appendix

\section*{Appendix}

In this appendix, we provide supplementary material to further elaborate the construction, evaluation protocol, and empirical analysis of OmniTraffic. The appendix is organized as follows to help readers quickly locate specific details:

\begin{itemize}[leftmargin=*]
    \item Appendix~\ref{app:dataset_construction} details dataset construction and visual diversity across simulated and real scenes.
    \item Appendix~\ref{app:task_taxonomy} expands the VQA task hierarchy with category-level question examples.
    \item Appendix~\ref{app:benchmark_protocol} reports the evaluation protocol and model details.
    \item Appendix~\ref{app:additional_results} provides additional benchmark diagnostics beyond the main text.
    \item Appendix~\ref{app:sim2real_detail} discusses sim-to-real experimental settings and results.
    \item Appendix~\ref{app:qualitative_case_studies} presents representative qualitative cases across the task hierarchy.
    \item Appendix~\ref{app:discussion} discusses the broader implications of OmniTraffic.
\end{itemize}

\section{Dataset Construction Details}
\label{app:dataset_construction}

OmniTraffic's simulated data is not generated from abstract or randomly designed road layouts. Instead, each simulated intersection is reconstructed from a real-world counterpart so that the generated images preserve realistic lane geometry, approach topology, signal layouts, and camera-view relationships. This section expands the dataset construction details along four axes. We first show the multi-view observations enabled by the real-aligned 3D scenes, then describe the traffic demand profiles used to create different temporal traffic states. We next present simulated special events and real-world visual variations that broaden the evaluation conditions. Finally, we describe the open 3D assets and custom rendering pipeline, which allows users to generate new views and extend OmniTraffic beyond the released image set.

\subsection{Multi-View Observations of Real-Aligned 3D Scenes}
\label{app:multi_views}

As discussed in Fig.~\ref{fig:intersections}, each reconstructed scene in OmniTraffic corresponds to a real-world intersection rather than a simulated toy layout. Figs.~\ref{fig_dataset_appendix_part1} illustrate the multi-view observations enabled by these 3D assets. Each row presents a top-down layout and directional inbound views, showing how the same intersection can be rendered from different camera placements. This design addresses the spatial perception gap in existing traffic benchmarks, which often rely on isolated frames or fixed viewpoints, whereas real traffic understanding requires correlating cross-view observations, inferring viewpoint-to-scene geometry, and grounding vehicles, lanes, and signals within a shared topology.

The 12 reconstructed intersections span Beijing, Chengdu, Tianjin, Massy, Yau Ma Tei, and Songdo, covering diverse geographic locations and road structures. Beihuan, Beishahe, Changjianglu, and Chenghannanlu are three-way intersections, while the remaining scenes are four-way intersections with varied lane organizations and approach layouts. Larger scenes such as Yongrunlu and Songdo provide broader road geometry and more complex traffic-state variations. This diversity allows OmniTraffic to evaluate MLLM generalization across viewpoints, intersection scale, topology, and regional road-layout patterns.

\begin{figure}[!htbp]
    \centering
    \subfloat[]{\includegraphics[width=1\textwidth]{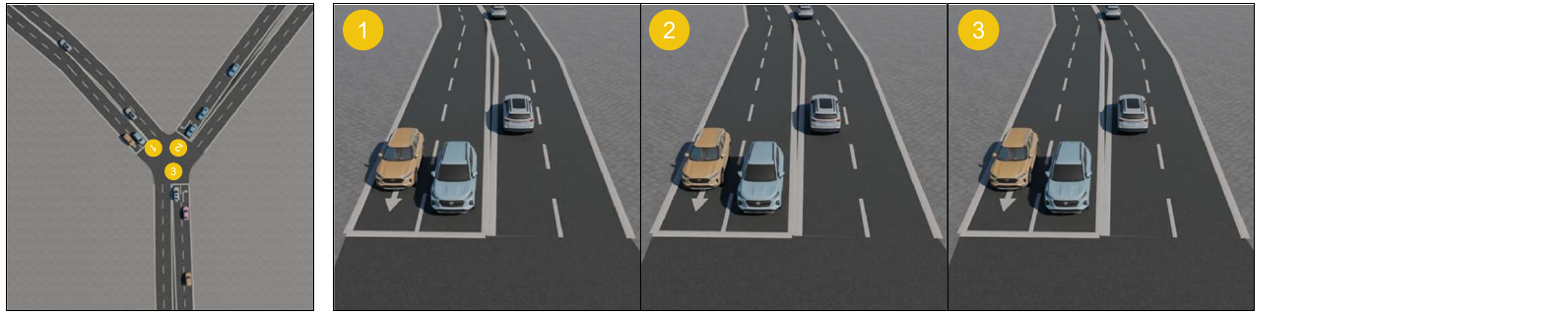}\label{fig_multiView_Beijing_Beihuan}}\\
    \caption{Multi-view observations of OmniTraffic intersections (Part 1). The row shows a top-down layout on the left and directional inbound views on the right. (a) Beihuan, Beijing.}
    \label{fig_dataset_appendix_part1}
\end{figure}

\begin{figure}[!htbp]
    \ContinuedFloat
    \centering
    \subfloat[]{\includegraphics[width=1\textwidth]{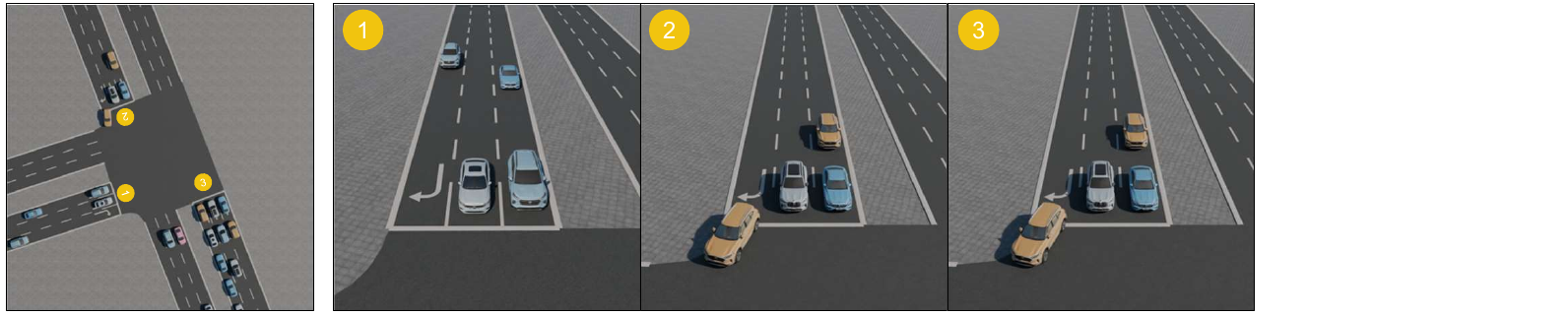}\label{fig_multiView_Beijing_Beishahe}}\\
    \subfloat[]{\includegraphics[width=1\textwidth]{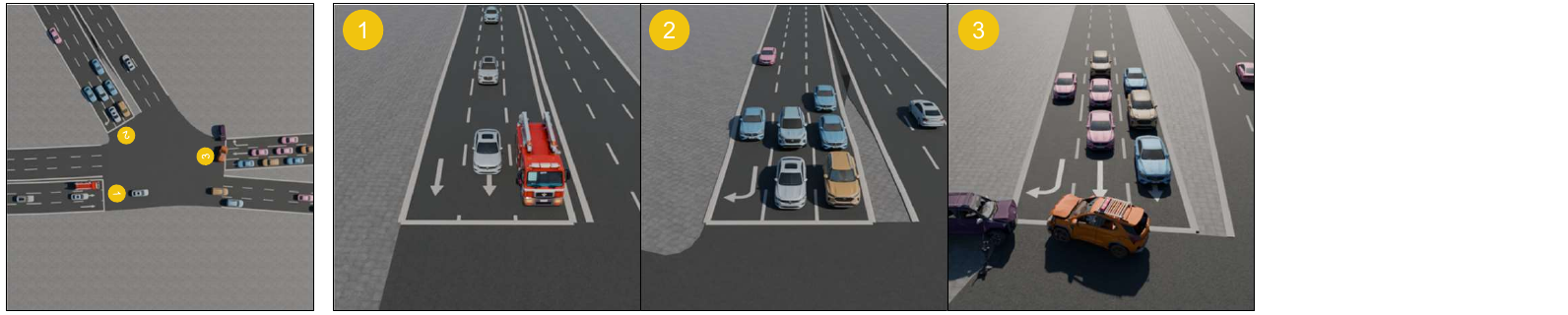}\label{fig_multiView_Beijing_Changjianglu}}\\
    \subfloat[]{\includegraphics[width=1\textwidth]{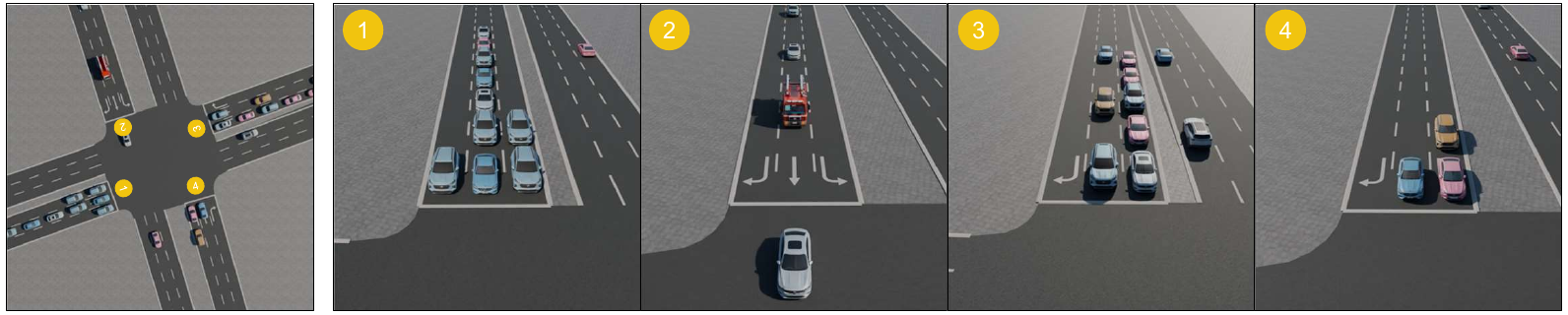}\label{fig_multiView_Beijing_Gaojiaoyuan}}\\
    \subfloat[]{\includegraphics[width=1\textwidth]{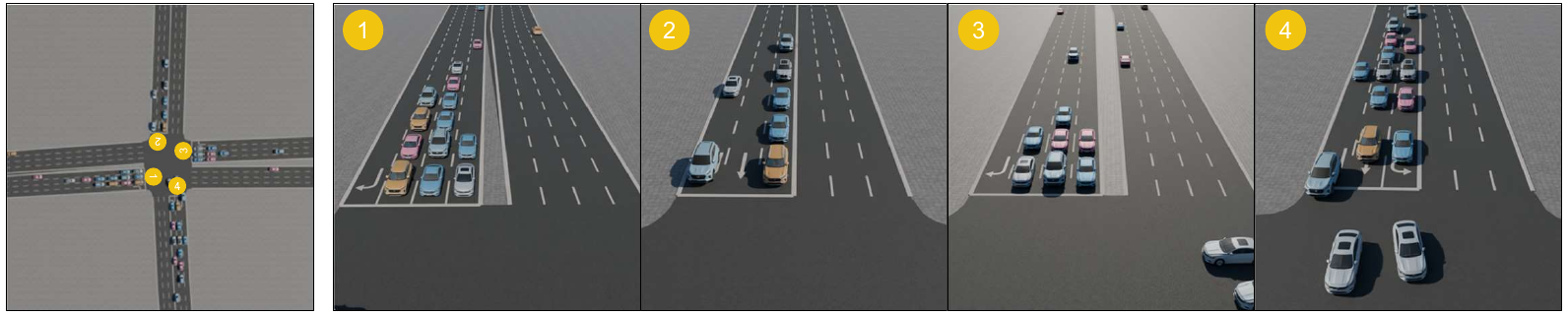}\label{fig_multiView_Beijing_Pinganli}}\\
    \subfloat[]{\includegraphics[width=1\textwidth]{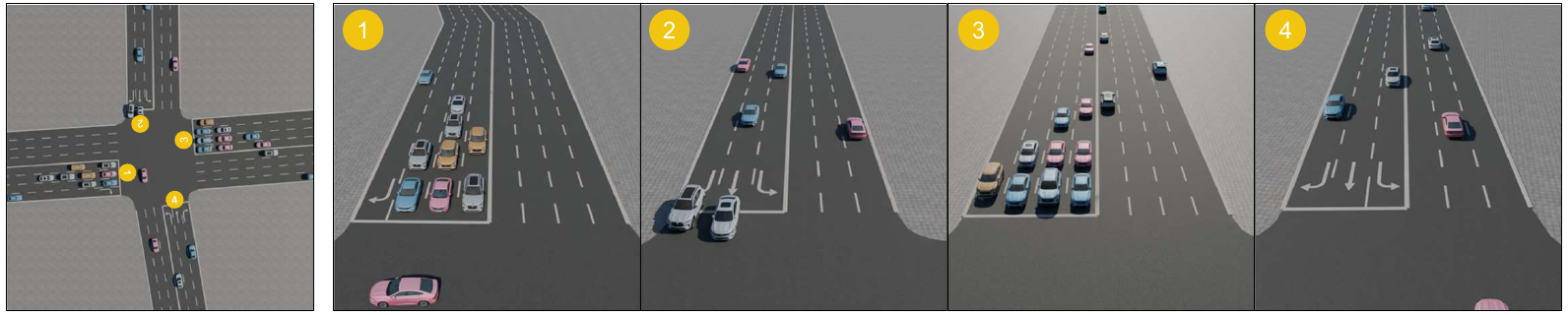}\label{fig_multiView_Beijing_Yongrunlu}}\\
    \subfloat[]{\includegraphics[width=1\textwidth]{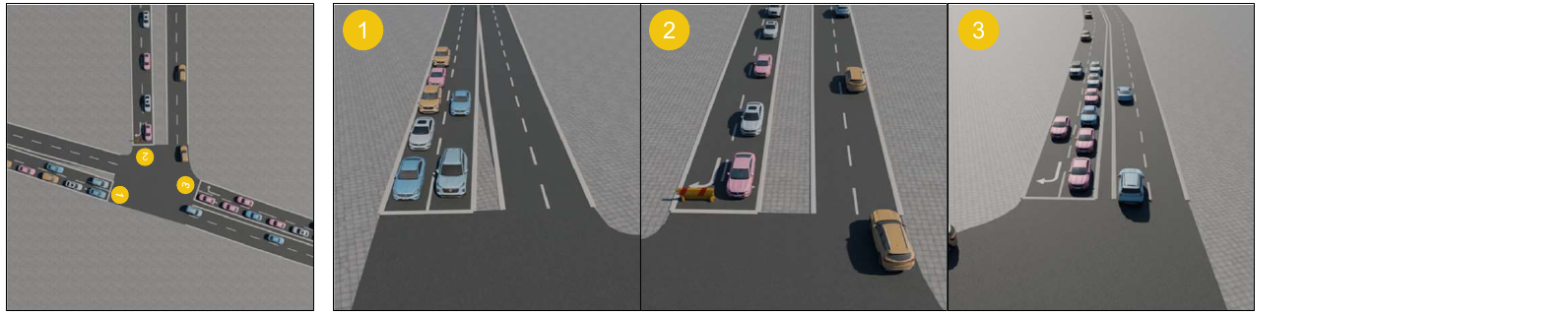}\label{fig_multiView_Chengdu_Chenghannanlu}}\\
    \caption{Multi-view observations of OmniTraffic intersections (Part 2). Each row shows a top-down layout and directional inbound views. (b) Beishahe, Beijing. (c) Changjianglu, Beijing. (d) Gaojiaoyuan, Beijing. (e) Pinganli, Beijing. (f) Yongrunlu, Beijing. (g) Chenghannanlu, Chengdu.}
    \label{fig_dataset_appendix_part2}
\end{figure}

\begin{figure}[!htbp]
    \ContinuedFloat
    \centering
    \subfloat[]{\includegraphics[width=1\textwidth]{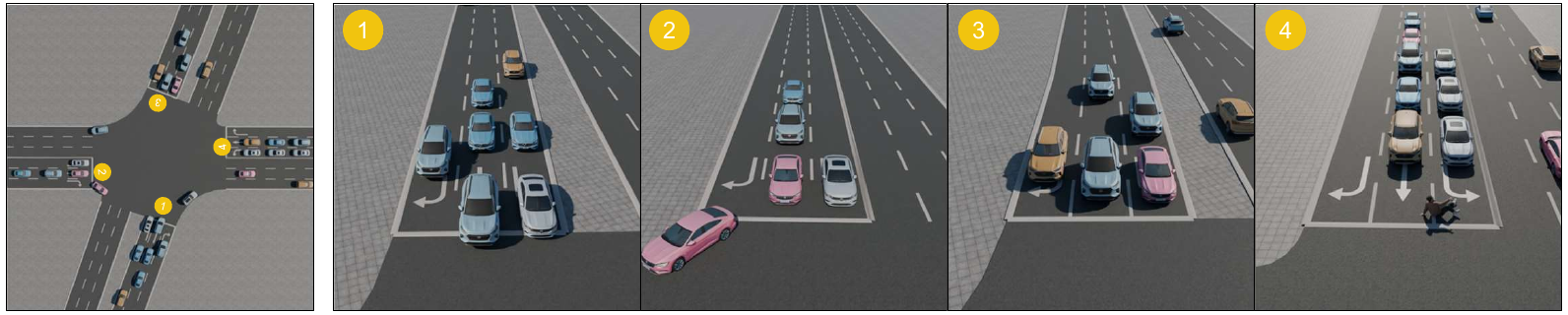}\label{fig_multiView_Chengdu_Guanghua}}\\
    \subfloat[]{\includegraphics[width=1\textwidth]{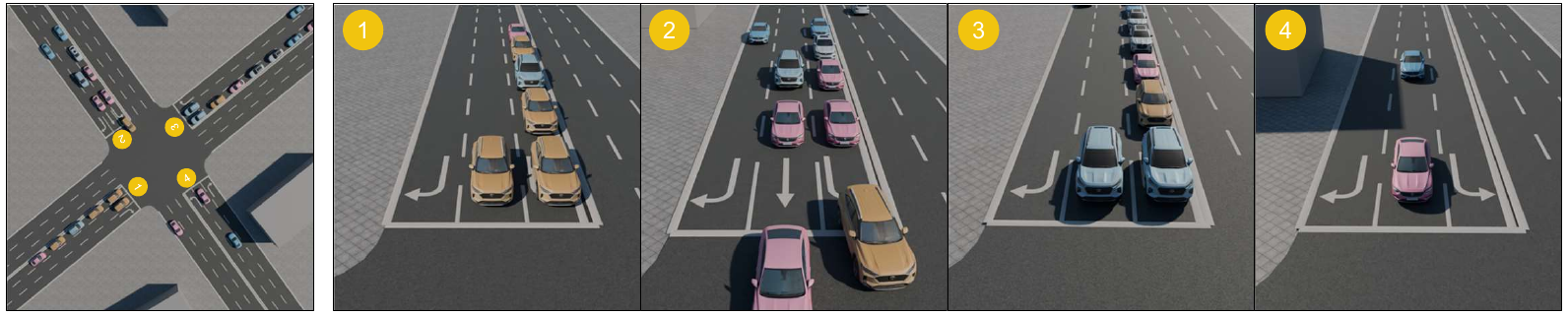}\label{fig_multiView_Tianjin_Zhijingdao}}\\
    \subfloat[]{\includegraphics[width=1\textwidth]{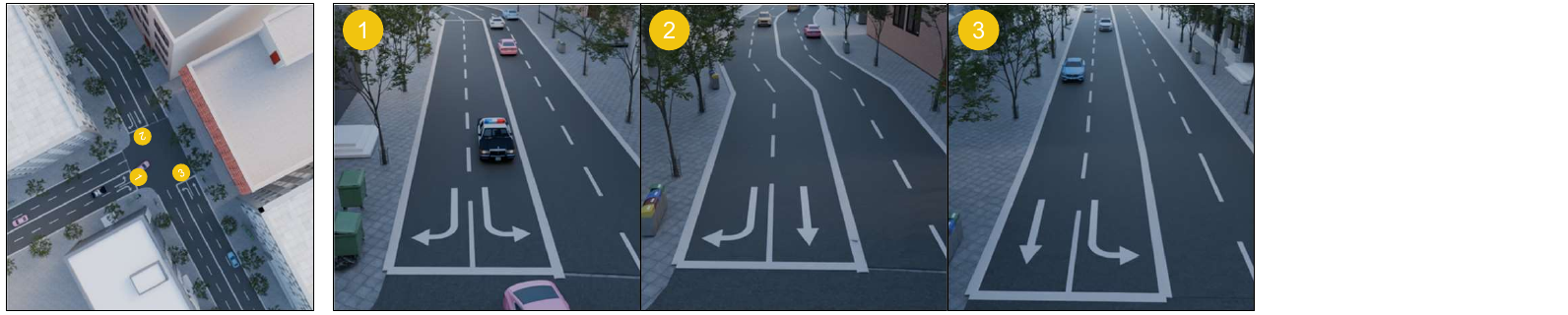}\label{fig_multiView_France_Massy}}\\
    \subfloat[]{\includegraphics[width=1\textwidth]{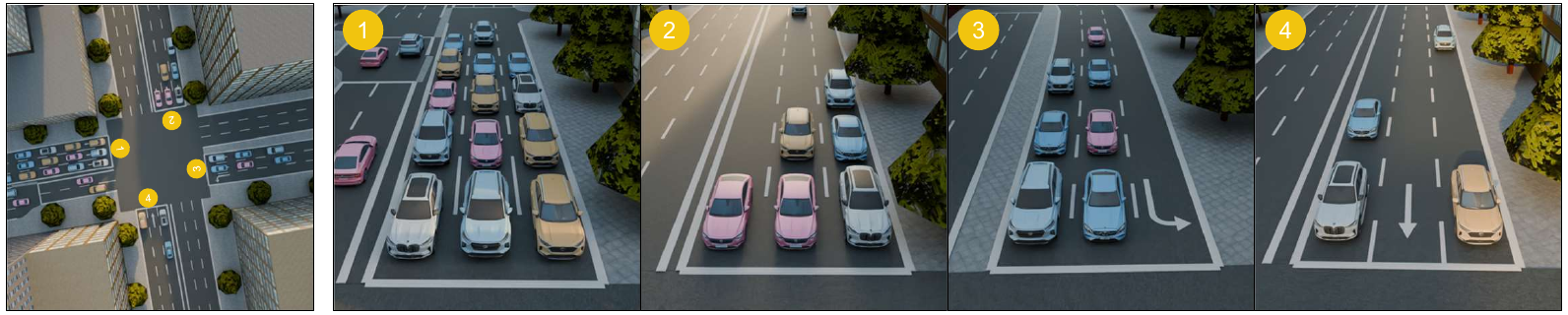}\label{fig_multiView_Hongkong_YMT}}\\
    \subfloat[]{\includegraphics[width=1\textwidth]{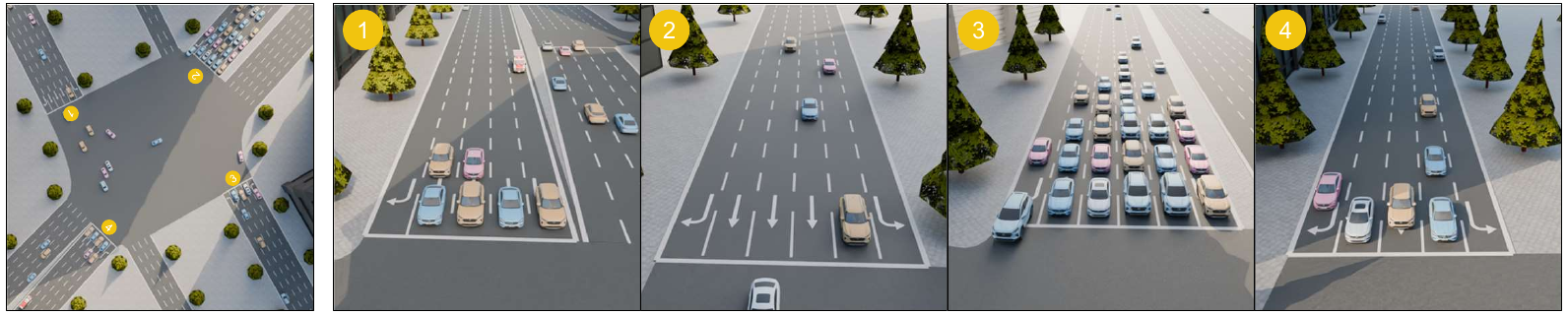}\label{fig_multiView_SouthKorea_Songdo}}\\
    \caption{Multi-view observations of OmniTraffic intersections (Part 3). Each row shows a top-down layout and directional inbound views. (h) Guanghua, Chengdu. (i) Zhijingdao, Tianjin. (j) Massy, France. (k) Yau Ma Tei, Hong Kong. (l) Songdo, South Korea.}
    \label{fig_dataset_appendix_part3}
\end{figure}

\subsection{Traffic Demand Profiles}
\label{app:traffic_demand_profiles}

To further diversify traffic states beyond intersection geometry and camera viewpoints, each real-aligned reconstructed scene is simulated under five traffic demand profiles: \emph{Low}, \emph{High}, \emph{Random}, \emph{Increase}, and \emph{Fluctuation}. These profiles vary both the overall traffic volume and its temporal evolution while keeping the underlying road topology fixed. This design ensures that OmniTraffic evaluates under different flow intensity.

Table~\ref{tab:traffic_flow_settings} reports the overall traffic-flow statistics for the 12 reconstructed scenes. Each entry gives the mean and standard deviation of vehicles per minute over a 10-minute simulation, providing a compact summary of the magnitude and variability induced by each demand profile. The table shows that the traffic profiles are scene-dependent: larger intersections such as Songdo generally support higher volumes, while smaller or three-way intersections exhibit lower overall demand under the same profile family.

\begin{table}[!htbp]
\centering
\caption{Overall traffic-flow statistics for the 12 reconstructed scenes. Values show mean $\pm$ standard deviation of vehicles per minute over a 10-minute simulation, computed from ten one-minute intervals.}
\label{tab:traffic_flow_settings}
\begin{tabular}{lccccc}
\toprule
\textbf{Scene} & \textbf{Low} & \textbf{High} & \textbf{Random} & \textbf{Incr.} & \textbf{Fluct.} \\
\midrule
Beijing Beihuan & $28.4 {\scriptstyle \pm 2.2}$ & $52.6{\scriptstyle \pm 3.2}$ & $45.6{\scriptstyle \pm 6.2}$ & $38.0{\scriptstyle \pm 10.3}$ & $44.4{\scriptstyle \pm 2.1}$ \\
Beijing Beishahe & $21.0{\scriptstyle \pm 1.9}$ & $67.4{\scriptstyle \pm 2.2}$ & $44.0{\scriptstyle \pm 11.1}$ & $47.6{\scriptstyle \pm 16.6}$ & $41.6{\scriptstyle \pm 12.3}$ \\
Beijing Changjianglu & $28.4{\scriptstyle \pm 2.2}$ & $58.4{\scriptstyle \pm 2.2}$ & $41.2{\scriptstyle \pm 10.2}$ & $47.8{\scriptstyle \pm 14.6}$ & $47.4{\scriptstyle \pm 21.4}$ \\
Beijing Gaojiaoyuan & $28.4{\scriptstyle \pm 1.5}$ & $70.0{\scriptstyle \pm 2.5}$ & $56.2{\scriptstyle \pm 8.4}$ & $54.6{\scriptstyle \pm 18.7}$ & $45.8{\scriptstyle \pm 15.6}$ \\
Beijing Pinganli & $34.0{\scriptstyle \pm 1.9}$ & $79.0{\scriptstyle \pm 2.5}$ & $47.4{\scriptstyle \pm 13.6}$ & $57.2{\scriptstyle \pm 16.2}$ & $54.6{\scriptstyle \pm 18.8}$ \\
Beijing Yongrunlu & $35.8{\scriptstyle \pm 2.5}$ & $77.0{\scriptstyle \pm 2.5}$ & $55.0{\scriptstyle \pm 12.7}$ & $53.4{\scriptstyle \pm 20.0}$ & $39.8{\scriptstyle \pm 24.4}$ \\
Chengdu Chenghannanlu & $20.2{\scriptstyle \pm 1.7}$ & $49.0{\scriptstyle \pm 1.3}$ & $32.2{\scriptstyle \pm 7.5}$ & $33.8{\scriptstyle \pm 14.9}$ & $31.8{\scriptstyle \pm 10.6}$ \\
Chengdu Guanghua & $29.6{\scriptstyle \pm 2.2}$ & $67.8{\scriptstyle \pm 2.9}$ & $46.8{\scriptstyle \pm 4.7}$ & $55.0{\scriptstyle \pm 20.4}$ & $54.6{\scriptstyle \pm 6.0}$ \\
France Massy & $20.4{\scriptstyle \pm 1.4}$ & $37.8{\scriptstyle \pm 7.0}$ & $42.0{\scriptstyle \pm 7.2}$ & $38.0{\scriptstyle \pm 8.3}$ & $37.8{\scriptstyle \pm 14.8}$ \\
Hongkong YMT & $30.8{\scriptstyle \pm 1.0}$ & $66.2{\scriptstyle \pm 2.4}$ & $44.6{\scriptstyle \pm 6.6}$ & $44.6{\scriptstyle \pm 11.1}$ & $40.4{\scriptstyle \pm 6.9}$ \\
SouthKorea Songdo & $50.0{\scriptstyle \pm 4.0}$ & $94.8{\scriptstyle \pm 3.6}$ & $87.2{\scriptstyle \pm 13.5}$ & $73.6{\scriptstyle \pm 24.6}$ & $64.2{\scriptstyle \pm 18.0}$ \\
Tianjin zhijingdao & $39.2{\scriptstyle \pm 1.5}$ & $70.6{\scriptstyle \pm 4.1}$ & $61.0{\scriptstyle \pm 8.7}$ & $50.6{\scriptstyle \pm 19.4}$ & $44.8{\scriptstyle \pm 13.0}$ \\
\bottomrule
\end{tabular}
\end{table}

\begin{figure}[!htbp]
  \centering
  \includegraphics[width=1\linewidth]{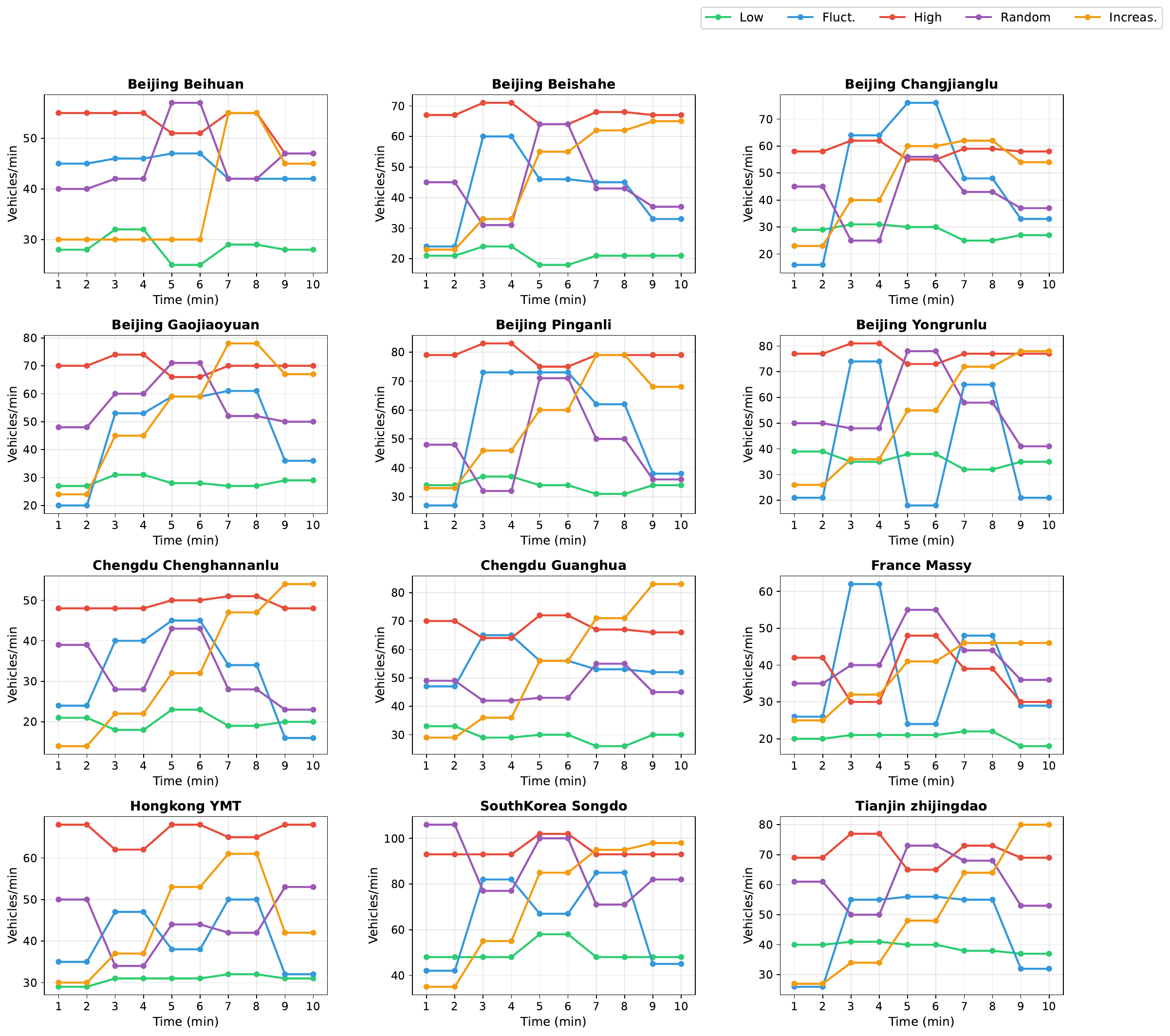}
  \caption{Temporal traffic demand profiles across the 12 real-aligned reconstructed OmniTraffic scenes. Each scene contains five curves corresponding to Low, High, Random, Increase, and Fluctuation demand profiles over ten one-minute intervals.}
  \label{fig:traffic_demand_profiles}
\end{figure}

Fig.~\ref{fig:traffic_demand_profiles} complements the table by showing how traffic demand changes over time in each reconstructed scene. The five curves correspond to the \emph{Low}, \emph{High}, \emph{Random}, \emph{Increase}, and \emph{Fluctuation} profiles across the ten one-minute intervals. This visualization shows that OmniTraffic controls not only the average traffic volume, but also the temporal pattern of demand, enabling questions that depend on queue evolution, temporal ordering, and signal-phase reasoning.

\subsection{Special Events and Real-World Visual Variations}
\label{app:special_conditions}

\begin{figure}[!htbp]
  \centering
  \includegraphics[width=1\linewidth]{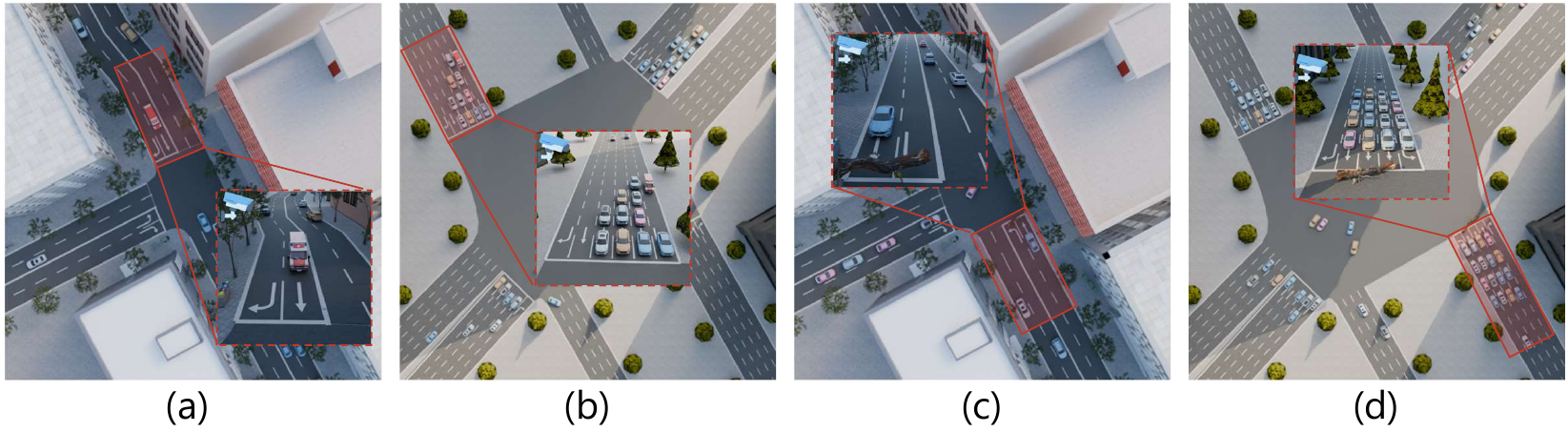}
  \caption{Representative special cases in simulated OmniTraffic scenes. (a)--(b) show emergency vehicles, including ambulances, while (c)--(d) show road-blocking events caused by fallen trees.}
  \label{fig_sim_special_cases_appendix}
\end{figure}

In addition to varying viewpoints and traffic demand, OmniTraffic includes controllable semantic events in simulation. Fig.~\ref{fig_sim_special_cases_appendix} shows representative simulated examples, including emergency vehicles and road-blocking events caused by fallen trees. These cases are designed to test whether MLLMs can recognize traffic-relevant entities and events that may alter local traffic states, lane availability, or signal-phase reasoning, rather than relying only on ordinary vehicle-flow patterns.

\begin{figure}[!htbp]
  \centering
  \includegraphics[width=1\linewidth]{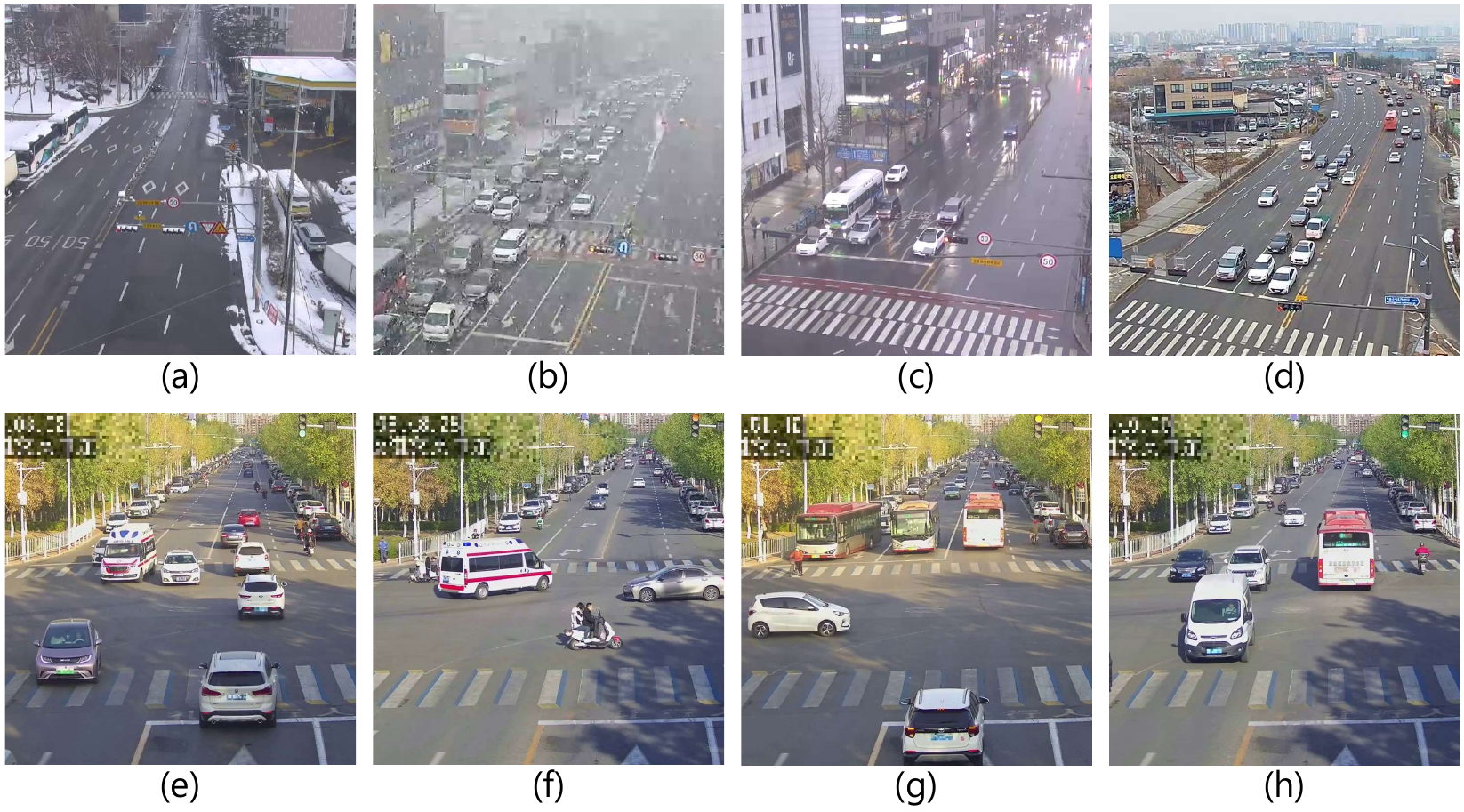}
  \caption{Representative weather conditions and special vehicles in real-world OmniTraffic footage. (a)--(b) show snowy scenes, (c) shows a rainy scene, (d) shows a cloudy scene, (e)--(f) show ambulances, and (g)--(h) show buses.}
  \label{fig_real_special_cases_appendix}
\end{figure}

Real-world footage further complements the simulated scenes by exposing models to visual conditions that are difficult to fully specify in simulation. Fig.~\ref{fig_real_special_cases_appendix} shows examples from real surveillance videos, including snowy, rainy, and cloudy weather, as well as emergency vehicles and buses. These examples help characterize the appearance gap considered in the sim-to-real experiments: models must handle not only structured traffic semantics, but also weather, visibility, and camera-domain variation.

\subsection{Open 3D Assets and Custom Rendering Pipeline}
\label{app:open_rendering_pipeline}

\begin{figure}[!htbp]
  \centering
  \includegraphics[width=1\linewidth]{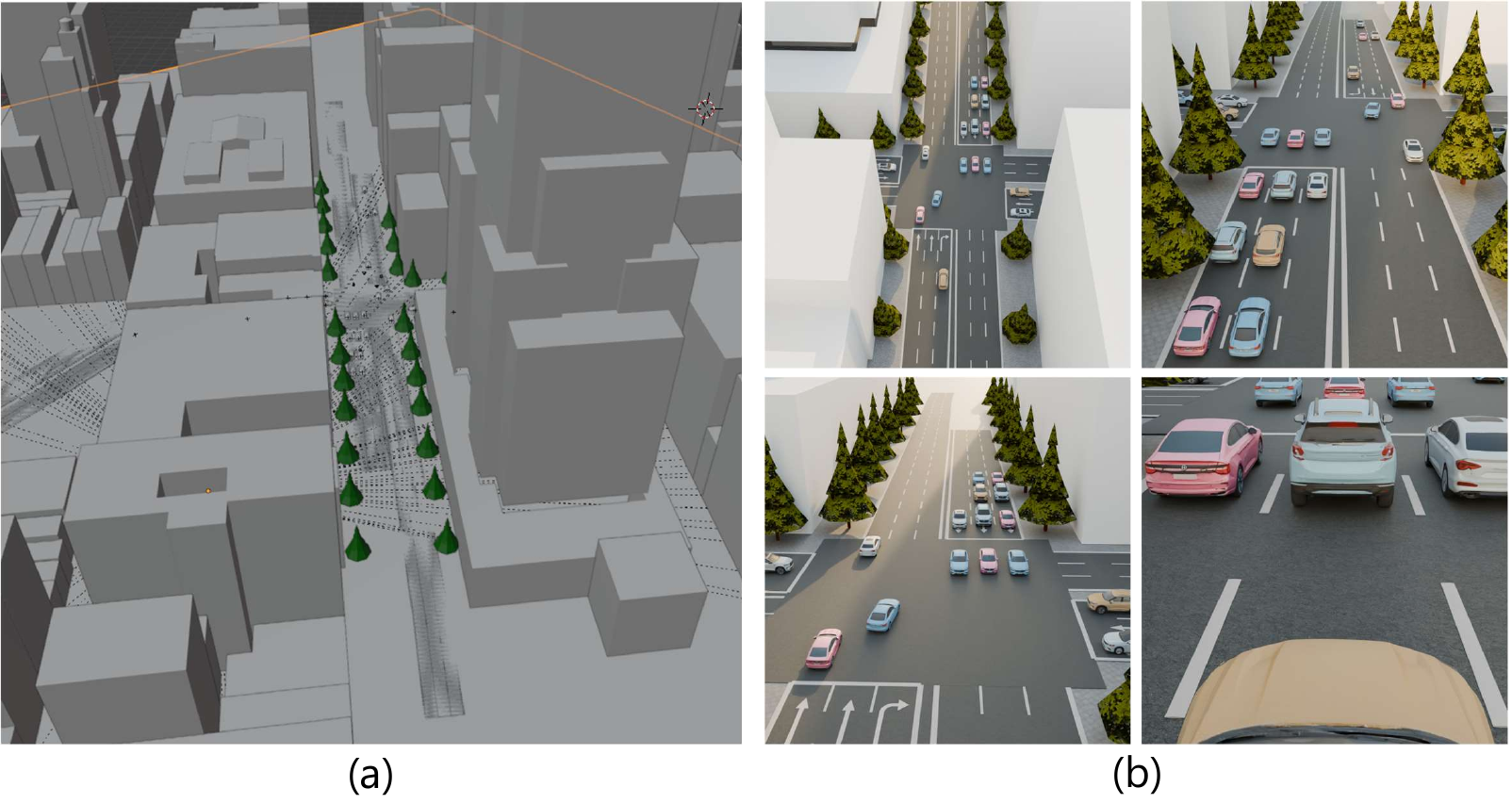}
  \caption{Open 3D assets and customizable rendering pipeline in OmniTraffic. (a) Reconstructed 3D intersection assets released with the benchmark. (b) Custom camera rendering enables the same traffic event to be captured from different viewpoints, including top-down, roadside intersection-camera, and autonomous-driving-style views.}
  \label{fig_open_rendering_pipeline}
\end{figure}

OmniTraffic is designed as an extensible scene resource rather than a fixed collection of released images. As shown in Fig.~\ref{fig_open_rendering_pipeline}(a), we release the reconstructed 3D intersection assets used to generate the simulated scenes. These assets preserve intersection geometry, lane topology, traffic-signal layouts, and scene context, enabling users to inspect, modify, and reuse the underlying environments instead of treating the benchmark as a closed set of images.

Together with the 3D assets, we will release the rendering code that controls camera placement and image generation. Fig.~\ref{fig_open_rendering_pipeline}(b) illustrates how the same traffic event can be rendered from different viewpoints, including top-down views, roadside intersection-camera views, and autonomous-driving-style views. This flexibility allows users to create additional images by changing camera angles, traffic demand profiles, special vehicles, special events, and visual conditions. The released pipeline therefore supports community-driven expansion of OmniTraffic and makes it possible to construct new VQA samples without manually rebuilding scene geometry or annotation logic.

\section{VQA Task Taxonomy and Question Examples}
\label{app:task_taxonomy}

The main text summarizes OmniTraffic's three-level task hierarchy in Table~\ref{tab:task_hierarchy}. This section further supplements the hierarchy with representative VQA examples covering all task categories. Each example includes the visual input, a natural-language question, four answer options, and the ground-truth answer, thereby illustrating the progression from single-image topology grounding to multi-view and temporal reasoning, and further to phase-level decision support. Table~\ref{tab:app_vqa_examples} maps each category to its corresponding example figure and reports the question used in that example. In the released benchmark, variables such as direction, lane index, timestep, image index, BEV option, and phase definition are instantiated from the corresponding simulation metadata or real-world annotations.

\begin{table}[!htbp]
\centering
\small
\caption{Representative VQA questions for each OmniTraffic category. Each row links to an example figure and lists the concrete question and answer choices shown in that example.}
\label{tab:app_vqa_examples}
\renewcommand{\arraystretch}{1.16}
\begin{tabularx}{\textwidth}{@{}p{0.1\textwidth}p{0.07\textwidth}X@{}}
\toprule
\textbf{Category} & \textbf{Example} & \textbf{Question and choices shown in the VQA example} \\
\midrule
\rowcolor{l1color!30}
\multicolumn{3}{@{}l}{\textbf{\textit{L1: Single-image perception examples}}} \\ \addlinespace[2pt]
Veh.~Count & Fig.~\ref{fig:app_vqa_veh_count} & Q: How many vehicles are there in incoming lane 2? Choices: (A) 1 (B) 2 (C) 3 (D) 4. \\
Spe.~Veh. & Fig.~\ref{fig:app_vqa_spe_veh} & Q: Which lane is the emergency vehicles located in (from left to right, lane index starts from 1)? Choices: (A) Lane 1 (B) Lane 2 (C) Lane 3 (D) Lane 4. \\
Spe.~Evt. & Fig.~\ref{fig:app_vqa_spe_evt} & Q: What types of traffic accidents or obstructions are visible in the image? Choices: (A) Safety Barrier (B) Vehicle Collision (C) Fallen Tree (D) None. \\
Road~Infra. & Fig.~\ref{fig:app_vqa_road_infra} & Q: What traffic movement is the first incoming lane from the left designated for? Choices: (A) left turn (B) straight (C) right turn (D) U-turn. \\
Scene~Attr. & Fig.~\ref{fig:app_vqa_scene_attr_examples} & Q: What is the weather condition in the image? Choices: (A) sunny (B) cloudy (C) rainy (D) snowy. \\
\midrule
\rowcolor{l2color!30}
\multicolumn{3}{@{}l}{\textbf{\textit{L2: Multi-image reasoning examples}}} \\ \addlinespace[2pt]
View~Comp. & Fig.~\ref{fig:app_vqa_view_comp} & Q: Which image has the most vehicles in incoming lanes? Choices: (A) Image 1 (B) Image 2 (C) Image 3 (D) Image 4. \\
View~Loc. & Fig.~\ref{fig:app_vqa_view_loc} & Q: In which image(s) are emergency vehicles visible? Choices: (A) Image 1 (B) Images 2 (C) Image 3 only (D) Image 4. \\
View-BEV & Fig.~\ref{fig:app_vqa_view_bev} & Q: Given the BEV (bird's-eye view), which directional view does NOT correspond to this BEV? Choices: (A)--(D) candidate view panels. \\
Temp.~Reas. & Fig.~\ref{fig:app_vqa_temp_reas} & Q: How does the vehicle queue change across the four frames? Choices: (A) increasing (B) decreasing (C) unchanged (D) first decreases then increases. \\
\midrule
\rowcolor{l3color!30}
\multicolumn{3}{@{}l}{\textbf{\textit{L3: Phase-level decision-support examples}}} \\ \addlinespace[2pt]
Phase~Ana. & Fig.~\ref{fig:app_vqa_phase_ana} & Q: Which traffic phase is affected by traffic accidents or obstructions? Choices: (A) Phase 1 (B) Phase 2 (C) Phase 3 (D) Phase 4. \\
Phase~Dec. & Fig.~\ref{fig:app_vqa_phase_dec} & Q: What is the optimal next green phase? Choices: (A) Phase 1 (B) Phase 2 (C) Phase 3 (D) Phase 4. \\
\bottomrule
\end{tabularx}
\end{table}

\subsection{Level 1: Perception}
\label{app:l1_examples}

\begin{figure}[!htbp]
    \centering
    \subfloat[Vehicle Counting.\label{fig:app_vqa_veh_count}]{
        \includegraphics[width=0.47\linewidth]{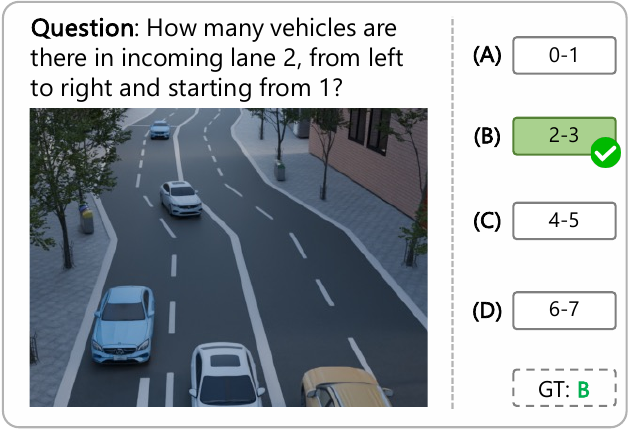}
    }
    \hfill
    \subfloat[Special Vehicle Recognition.\label{fig:app_vqa_spe_veh}]{
        \includegraphics[width=0.47\linewidth]{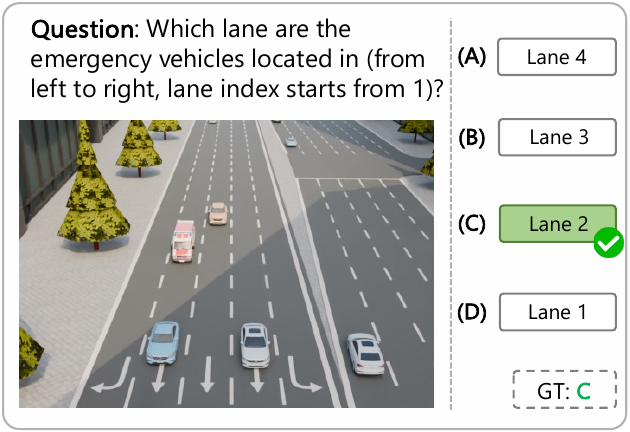}
    }
    \caption{Representative VQA examples for Vehicle Counting and Special Vehicle Recognition.}
    \label{fig:app_vqa_l1_count_special_vehicle_examples}
\end{figure}

\begin{figure}[!htbp]
    \centering
    \subfloat[Special Event Recognition.\label{fig:app_vqa_spe_evt}]{
        \includegraphics[width=0.47\linewidth]{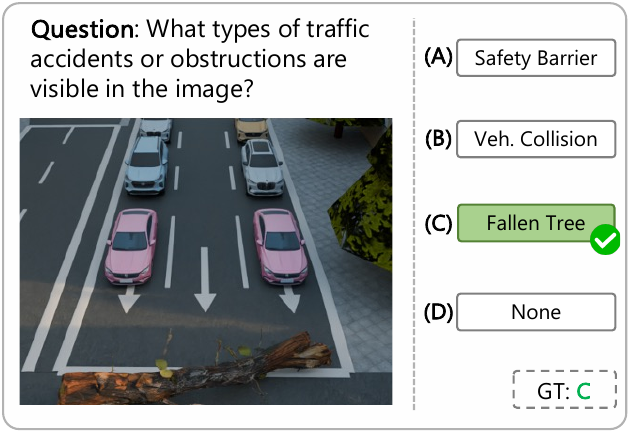}
    }
    \hfill
    \subfloat[Road Infrastructure.\label{fig:app_vqa_road_infra}]{
        \includegraphics[width=0.47\linewidth]{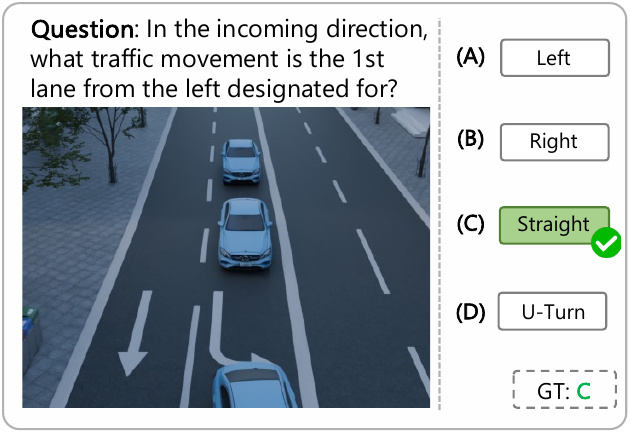}
    }
    \caption{Representative VQA examples for Special Event Recognition and Road Infrastructure.}
    \label{fig:app_vqa_l1_event_infrastructure_examples}
\end{figure}

\begin{figure}[!htbp]
    \centering
    \subfloat[Traffic-light color.\label{fig:app_vqa_scene_attr_signal}]{
        \includegraphics[width=0.47\linewidth]{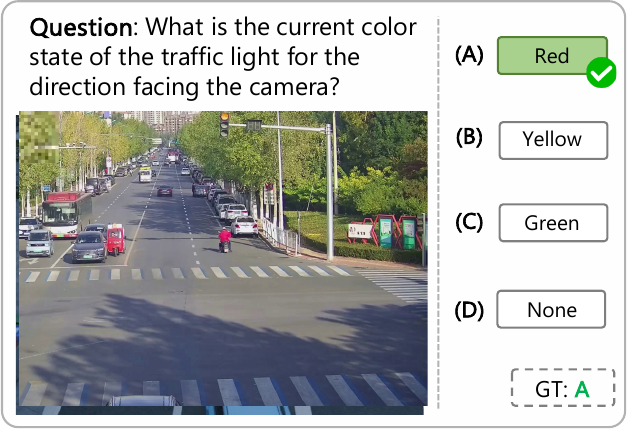}
    }
    \hfill
    \subfloat[Weather condition.\label{fig:app_vqa_scene_attr_weather}]{
        \includegraphics[width=0.47\linewidth]{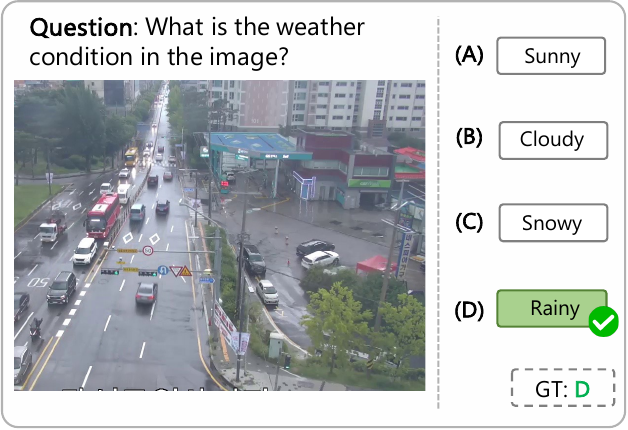}
    }
    \caption{Representative VQA examples for Scene Attribute. The examples ask the model to recognize the traffic-light color state and the weather condition from real-world surveillance frames.}
    \label{fig:app_vqa_scene_attr_examples}
\end{figure}

Level 1 evaluates whether a model can convert a single traffic image into structured facts grounded in road topology. Although the input is a single frame, the answer often depends on visibility, direction, lane indexing, or lane function rather than generic object recognition alone. Fig.~\ref{fig:app_vqa_veh_count}--Fig.~\ref{fig:app_vqa_scene_attr_examples} illustrate the five L1 categories with single-image VQA examples.

\subsubsection{Vehicle Counting (Veh.~Count)}
\label{app:l1_vehicle_count_examples}

Vehicle-counting questions evaluate whether the model can count visible vehicles at different levels of granularity. Beyond conventional whole-image vehicle counting, OmniTraffic introduces lane-specific counting tasks that query the number of vehicles in a particular lane or movement direction. This design is closer to real-world traffic monitoring, where effective perception requires not only a global estimate of traffic volume but also a fine-grained understanding of how vehicles are distributed across lanes. A representative VQA example should therefore present a single traffic image with labeled incoming and outgoing lanes, followed by a lane-grounded counting question such as:
\begin{enumerate}[leftmargin=*,label=\arabic*.]
    \item How many vehicles are there in total on the incoming and outgoing lanes, considering visibility?
    \item How many vehicles are there in the incoming direction, considering visibility?
    \item How many vehicles are there in the outgoing direction, considering visibility?
    \item How many vehicles are there in incoming lane 2, from left to right and starting from 1?
\end{enumerate}
The last template is instantiated with different directions and lane indices, which tests whether models can associate vehicles with a specific lane rather than only estimate scene-level traffic density. The answer options are numeric counts sampled from plausible nearby values.

\subsubsection{Special Vehicle Recognition (Spe.~Veh.)}
\label{app:l1_special_vehicle_examples}

Special-vehicle questions target traffic-relevant vehicle classes such as police cars, ambulances, and fire trucks. A representative VQA example should highlight the special vehicle only subtly, if at all, so that the example remains faithful to the benchmark setting. The questions progress from existence detection to type recognition and lane-level localization:
\begin{enumerate}[leftmargin=*,label=\arabic*.]
    \item Does the image contain any emergency vehicles such as police cars, ambulances, or fire trucks?
    \item What type of emergency vehicle is shown in the image, and where is it located?
    \item Which lane are the emergency vehicles located in, from left to right and starting from index 1?
\end{enumerate}

\subsubsection{Special Event Recognition (Spe.~Evt.)}
\label{app:l1_special_event_examples}

Special-event questions focus on rare but traffic-critical conditions that can change lane availability or downstream signal decisions. A representative VQA example should show an obstruction or accident in a specific lane and use options that differ by event type or affected lane:
\begin{enumerate}[leftmargin=*,label=\arabic*.]
    \item Is there any traffic accident or obstruction visible in the image?
    \item What types of traffic accidents or obstructions are visible in the image?
    \item Which lanes are affected by the traffic accident or obstruction, from left to right and starting from lane index 1?
\end{enumerate}

\subsubsection{Road Infrastructure (Road Infra.)}
\label{app:l1_road_infra_examples}

Road-infrastructure questions require models to read the static structure of the intersection, including the number of lanes and their permitted movements. A representative VQA example should use a clear image where lane markings or arrows are visible:
\begin{enumerate}[leftmargin=*,label=\arabic*.]
    \item How many incoming lanes are there in total?
    \item How many outgoing lanes are there in total?
    \item How many straight lanes are there in the incoming direction?
    \item In the incoming direction, what traffic movement is the third lane from the left designated for?
\end{enumerate}
The direction, movement type, and ordinal lane reference vary across samples, enabling evaluation of lane-topology grounding under different road layouts.

\subsubsection{Scene Attribute (Scene Attr.)}
\label{app:l1_scene_attribute_examples}

Scene-attribute questions are constructed from real-world footage. We use two representative examples for this category: one asks for the traffic-light color state, and the other asks for the weather condition. Both examples use real surveillance frames and target a visible attribute rather than the entire scene:
\begin{enumerate}[leftmargin=*,label=\arabic*.]
    \item Within the main zebra crossing directly in front of the camera, how many pedestrians are present?
    \item At the intersection, the traffic light head follows the standard vertical configuration: red on top, yellow in the middle, and green at the bottom. Based on which lamp is illuminated, what is the current color state of the traffic light for the direction facing the camera?
    \item What is the weather condition in the image?
\end{enumerate}

\subsection{Level 2: Spatiotemporal Reasoning}
\label{app:l2_examples}

\begin{figure}[!htbp]
    \centering
    \includegraphics[width=0.92\linewidth]{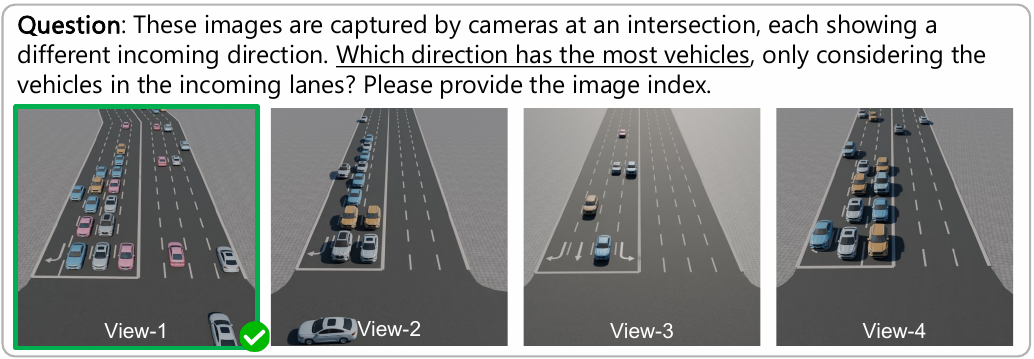}
    \caption{Representative VQA example for Multi-view Comparison. The example asks the model to compare incoming-lane vehicle counts across directional views.}
    \label{fig:app_vqa_view_comp}
\end{figure}

\begin{figure}[!htbp]
    \centering
    \includegraphics[width=0.92\linewidth]{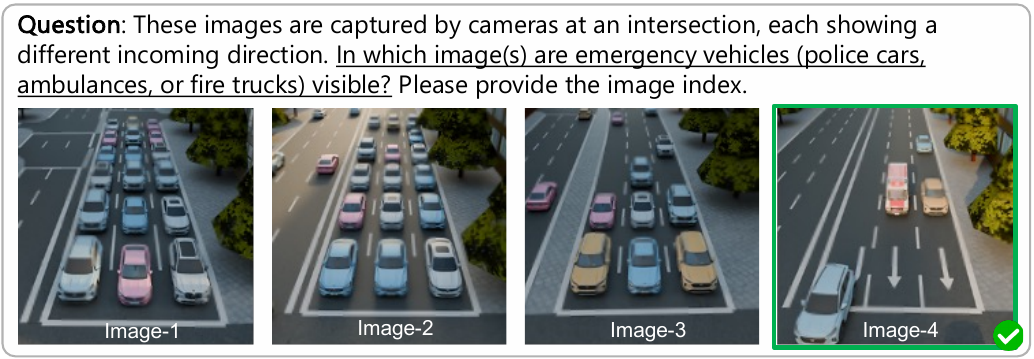}
    \caption{Representative VQA example for Multi-view Localization. The example asks the model to find which directional view(s) contain a traffic-relevant target.}
    \label{fig:app_vqa_view_loc}
\end{figure}

\begin{figure}[!htbp]
    \centering
    \includegraphics[width=0.92\linewidth]{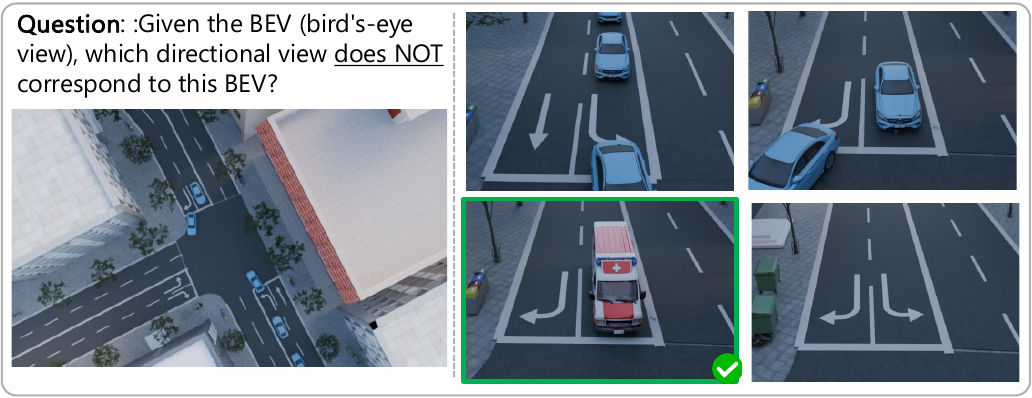}
    \caption{Representative VQA example for View-BEV Mapping. The example asks the model to align a perspective view with the corresponding bird's-eye-view geometry.}
    \label{fig:app_vqa_view_bev}
\end{figure}

\begin{figure}[!htbp]
    \centering
    \includegraphics[width=0.92\linewidth]{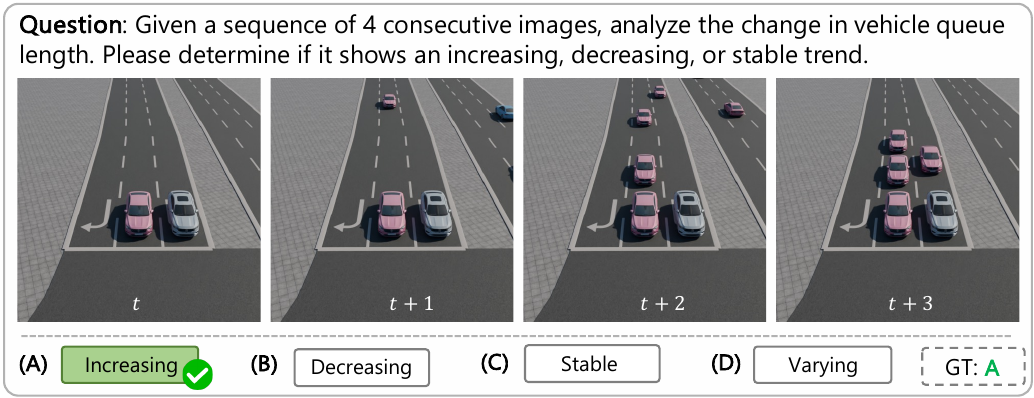}
    \caption{Representative VQA example for Temporal Reasoning. The example asks the model to infer queue evolution from multiple frames.}
    \label{fig:app_vqa_temp_reas}
\end{figure}

Level 2 extends the input from a single image to a set of images. The model must establish correspondences across views, BEV layouts, or timesteps before answering. These tasks are designed to distinguish models that recognize local content from models that can reason over the shared intersection state. Fig.~\ref{fig:app_vqa_view_comp}--Fig.~\ref{fig:app_vqa_temp_reas} illustrate the four L2 categories with multi-image VQA examples.

\subsubsection{Multi-view Comparison (View~Comp.)}
\label{app:l2_view_comp_examples}

Multi-view comparison questions present multiple synchronized directional views from the same intersection. A representative VQA example should arrange four directional images in a grid and ask for the image index with the largest incoming queue:
\begin{enumerate}[leftmargin=*,label=\arabic*.]
    \item These images are captured by cameras at an intersection, each showing a different incoming direction. Which direction has the most vehicles, only considering the vehicles in the incoming lanes? Please provide the image index.
\end{enumerate}

\subsubsection{Multi-view Localization (View~Loc.)}
\label{app:l2_view_loc_examples}

Multi-view localization questions ask the model to search across views and return the image index or indices containing traffic-relevant targets. A representative VQA example should show multiple camera views where only a subset contains an emergency vehicle or special event:
\begin{enumerate}[leftmargin=*,label=\arabic*.]
    \item These images are captured by cameras at an intersection, each showing a different incoming direction. In which image(s) are emergency vehicles, such as police cars, ambulances, or fire trucks, visible? Please provide the image index.
    \item These images are captured by cameras at an intersection, each showing a different incoming direction. In which image(s) are special events visible? Please provide the image index.
\end{enumerate}

\subsubsection{View-BEV Mapping (View-BEV)}
\label{app:l2_view_bev_examples}

View-BEV mapping questions connect perspective observations with bird's-eye-view layouts. This category directly evaluates whether models can infer the geometric relation between directional camera views and the global intersection map. A representative VQA example should combine one BEV map and several perspective-view options, or one perspective image and several BEV options:
\begin{enumerate}[leftmargin=*,label=\arabic*.]
    \item Given a BEV, or bird's-eye view, determine the driving direction associated with the star-shaped marker. Please select the correct description from the provided options.
    \item Given the BEV, which directional view does not correspond to this BEV?
    \item Given the directional view, which BEV corresponds to this view?
\end{enumerate}

\subsubsection{Temporal Reasoning (Temp.~Reas.)}
\label{app:l2_temporal_examples}

Temporal-reasoning questions use frames from the same viewpoint at different timesteps. They evaluate ordering, interpolation, prediction, and traffic-state trend recognition. A representative VQA example should show a short ordered or partially ordered sequence from the same camera:
\begin{enumerate}[leftmargin=*,label=\arabic*.]
    \item Given two reference images from the start and end times, which of the following images occurred between these two time points?
    \item Among these images from the same viewpoint at different timesteps, which one occurred first?
    \item The image shows the vehicle queue situation in the same direction across four consecutive timesteps. Please determine whether the vehicle queue in this direction is increasing, decreasing, or remaining unchanged.
    \item Given two consecutive reference images from timesteps, which of the following images would occur next in the sequence?
\end{enumerate}

\subsection{Level 3: Decision Support}
\label{app:l3_examples}

\begin{figure}[!htbp]
    \centering
    \includegraphics[width=0.92\linewidth]{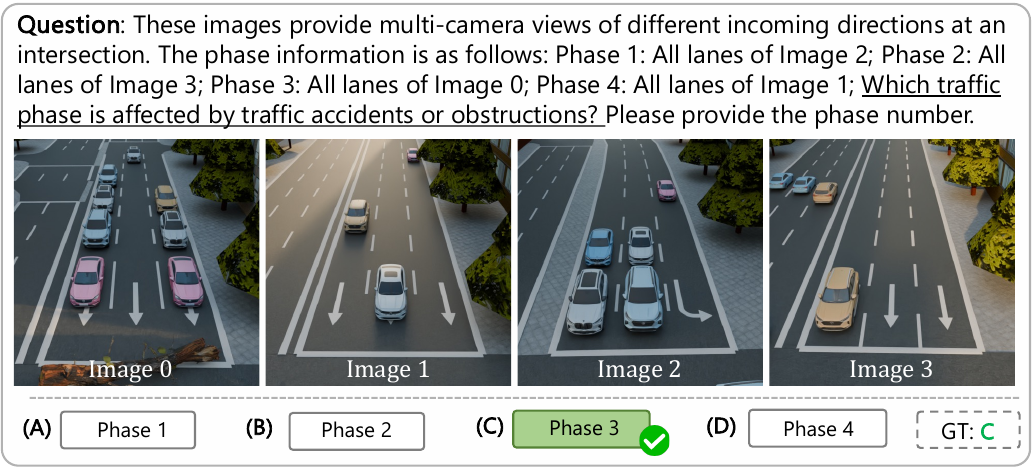}
    \caption{Representative VQA example for Signal Phase Analysis. The example asks the model to map emergency events to a traffic phase.}
    \label{fig:app_vqa_phase_ana}
\end{figure}

\begin{figure}[!htbp]
    \centering
    \includegraphics[width=0.92\linewidth]{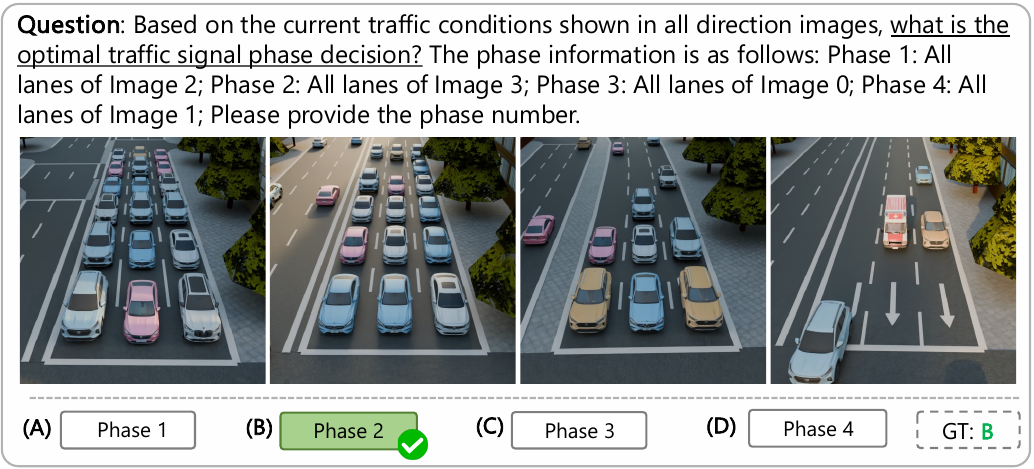}
    \caption{Representative VQA example for Signal Phase Decision. The example asks the model to recommend the next green phase from the current multi-view traffic condition.}
    \label{fig:app_vqa_phase_dec}
\end{figure}

Level 3 uses the perceptual and spatiotemporal facts from earlier levels as inputs to signal-phase reasoning. Each sample provides synchronized directional views together with phase definitions, such as the lane movements served by each phase. The model must map observed traffic states to the corresponding phases before selecting an answer. Fig.~\ref{fig:app_vqa_phase_ana} and Fig.~\ref{fig:app_vqa_phase_dec} illustrate the two L3 categories with phase-level decision-support examples.

\subsubsection{Signal Phase Analysis (Phase~Ana.)}
\label{app:l3_phase_analysis_examples}

Signal-phase analysis questions ask models to identify which phase is most relevant under the current traffic condition. The evidence may be ordinary queue length, special events, emergency vehicles, or the observed signal color. A representative VQA example should show all directional views and the phase table so that readers can see how visual evidence maps to a phase number:
\begin{enumerate}[leftmargin=*,label=\arabic*.]
    \item These images are captured by cameras at an intersection, each showing a different incoming direction. The phase information is provided. Which traffic phase has the most vehicles waiting? Please provide the phase number.
    \item These images are captured by cameras at an intersection, each showing a different incoming direction. The phase information is provided. Which traffic phase is affected by traffic accidents or obstructions? Please provide the phase number.
    \item These images are captured by cameras at an intersection, each showing a different incoming direction. The phase information is provided. Which traffic phase contains emergency vehicles, such as police cars, ambulances, or fire trucks? Please provide the phase number.
    \item These images are captured by cameras at an intersection, each showing a different incoming direction. The phase information is provided. Which traffic phase currently has a green light? Please provide the phase number.
\end{enumerate}

\subsubsection{Signal Phase Decision (Phase~Dec.)}
\label{app:l3_phase_decision_examples}

Signal-phase decision questions move from analysis to recommendation. Instead of asking the model to report an observed phase property, the model must choose the next green phase based on the complete multi-view traffic state and the available phase definitions. A representative VQA example should show the current traffic condition in all directions, the candidate phase definitions, and four phase-number options:
\begin{enumerate}[leftmargin=*,label=\arabic*.]
    \item Based on the current traffic conditions shown in all direction images, what is the optimal traffic signal phase decision? The phase information is provided. Please provide the phase number.
\end{enumerate}

\section{Evaluation Protocol and Model Details}
\label{app:benchmark_protocol}

The verified OmniTraffic benchmark is evaluated in a unified four-option multiple-choice format so that human and model results are directly comparable. Before evaluation, questions with ambiguous visual evidence, near-duplicate wording, or weak distractors are removed or replaced to reduce artifacts unrelated to traffic understanding. We therefore focus this appendix section on three protocol details that complement the main text: the web platform used for human evaluation, the deterministic model scoring procedure, and the full model identifiers used in the experiments.

\subsection{Human Evaluation Platform and Protocol}
\label{app:human_eval_interface}

\begin{figure}[!htbp]
    \centering
    \begin{minipage}{0.32\textwidth}
        \centering
        \includegraphics[width=\linewidth]{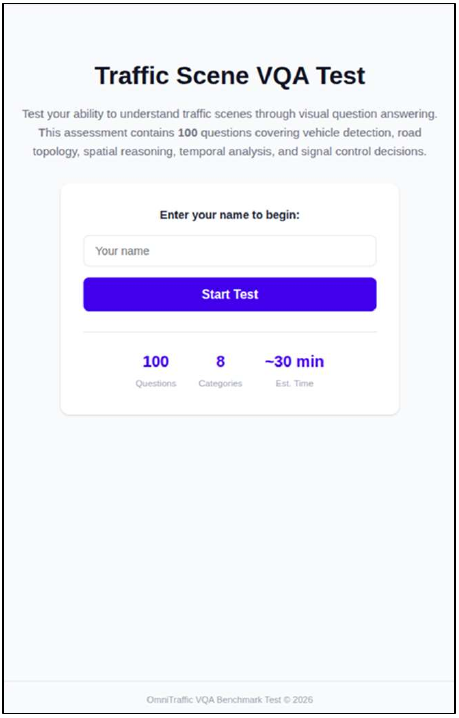}
        \centerline{(a) Login interface}
    \end{minipage}
    \hfill
    \begin{minipage}{0.32\textwidth}
        \centering
        \includegraphics[width=\linewidth]{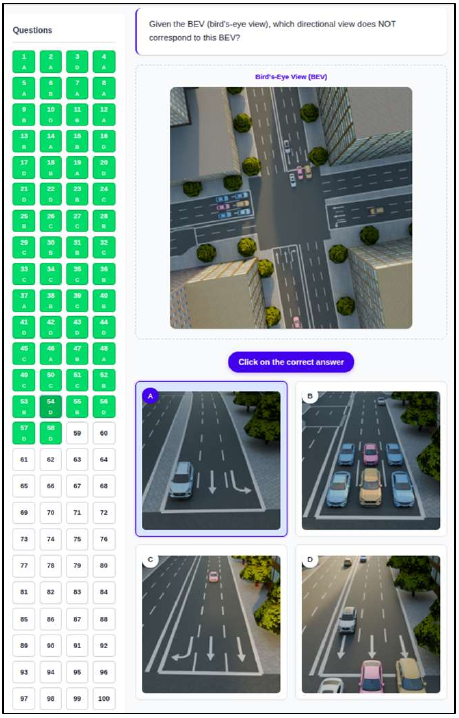}
        \centerline{(b) Task interface}
    \end{minipage}
    \hfill
    \begin{minipage}{0.32\textwidth}
        \centering
        \includegraphics[width=\linewidth]{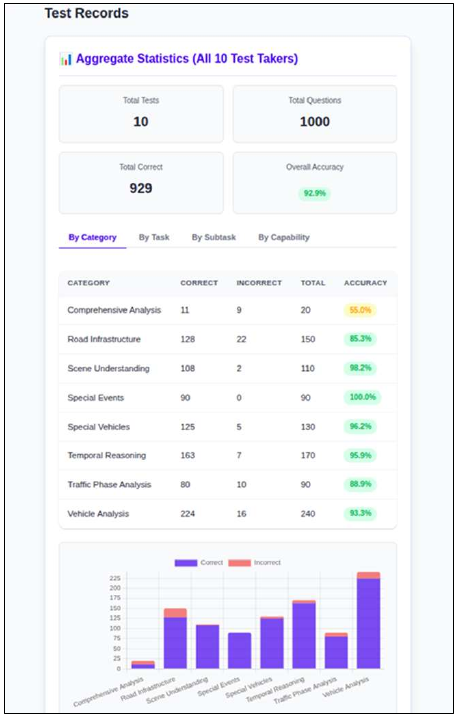}
        \centerline{(c) Score interface}
    \end{minipage}
    \caption{Screenshots of the web interface developed for human evaluation. (a) The login interface where users enter their usernames. (b) The task interface, where green indicators represent completed tasks. (c) The final score interface showing user statistics and scores after completion.}
    \label{fig_web_interface}
\end{figure}

Human evaluation is conducted through a web-based platform using the same multiple-choice benchmark format as model evaluation. Ten participants with no specialized traffic-engineering background complete the benchmark without external aids. As shown in Fig.~\ref{fig_web_interface}, the platform contains a login page, a task page, and a score page. The task page presents the visual input, question, and four candidate answers, while tracking which questions have been completed. After submission, the score page reports the participant's final statistics. Human accuracy is computed by exact match against the ground-truth option, allowing a direct comparison with MLLMs under the same scoring rule.

\subsection{Model Evaluation Protocol}
\label{app:model_eval_protocol}

For each benchmark item, the model receives the visual input, the question, and four answer options. The prompt asks the model to reason over the traffic scene and provide an answer in the option format, so model outputs may include explanatory text in addition to the selected choice. To make comparisons reproducible across providers, all evaluations use deterministic decoding whenever the interface allows it: temperature is set to $\tau = 0$, top-$p$ is kept at its default value of 1.0 when exposed by the API, frequency penalty is kept at its default value of 0, and the maximum output length is set to 500 generated tokens. No external tools, retrieval modules, or task-specific demonstrations are provided. Other provider-specific sampling parameters are left at their default values when they are not exposed consistently across model APIs.

\begin{figure}[!htbp]
    \centering
    \includegraphics[width=0.92\linewidth]{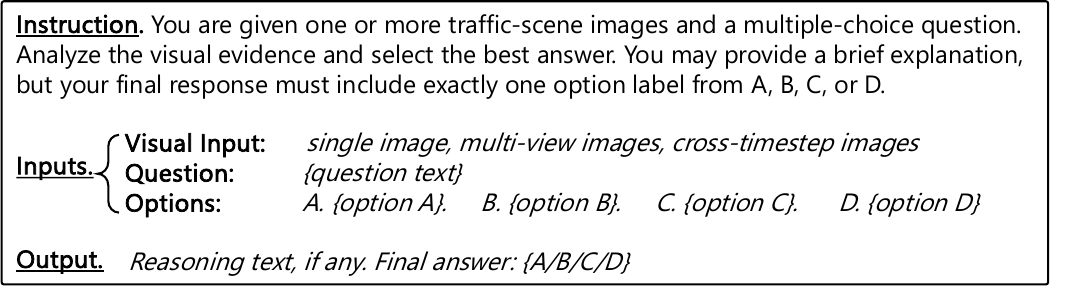}
    \caption{Prompt template used for model evaluation. The template permits explanatory reasoning while requiring an explicit option label for deterministic answer extraction.}
    \label{fig:prompt_template}
\end{figure}

Scoring is performed only on the extracted option label rather than on the explanatory text. We use a deterministic regular-expression parser to identify valid option labels in common answer formats, such as ``A'', ``(A)'', ``Option A'', or sentences containing an explicit final choice. If no valid label can be extracted, the same query is retried up to three times with the same visual input and question. A response is counted as invalid only if all three attempts fail to produce a parsable option, and invalid responses are scored as incorrect. Model accuracy is then reported as exact-match accuracy between the extracted option and the ground-truth label. The invalid-response rate in Appendix~\ref{app:response_reliability} separately reports this format-compliance failure mode, so it is not conflated with traffic-scene reasoning errors.

\subsection{Model Identifiers}
\label{app:model_identifiers}

Table~\ref{tab:model_identifiers} lists the full model identifiers corresponding to the shortened names used in the result tables. The shortened names are used only for readability and do not change the evaluated model variants.

\begin{table}[!htbp]
\centering
\caption{Full model identifiers for the abbreviated names used in the experiment tables.}
\label{tab:model_identifiers}
\begin{tabular}{ll}
\toprule
\textbf{Short name} & \textbf{Full identifier} \\
\midrule
GPT-4o & GPT-4o \\
GPT-5.2 & GPT-5.2 \\
Gemini-2.5-Pro & Gemini-2.5-Pro \\
Gemini-3-Pro & Gemini-3-Pro-Thinking \\
Claude-Sonnet-4.5 & Claude-Sonnet-4.5-Thinking \\
Doubao-1.5 & Doubao-1.5-Thinking-Vision-Pro-250428 \\
Grok-4 & Grok-4 \\
Qwen-VL-Max & Qwen-VL-Max-2025-04-08 \\
Qwen3-VL-Plus & Qwen3-VL-Plus \\
\midrule
InternVL & InternVL 3.5 \\
Qwen3-VL-235B & Qwen3-VL-235B-A22B-Thinking \\
\bottomrule
\end{tabular}
\end{table}

\section{Additional Benchmark Analysis}
\label{app:additional_results}

The main text argues that current MLLMs are relatively strong at salient visual recognition but weak at lane-level topology grounding, quantitative traffic-state estimation, and temporal reasoning. This section examines whether that interpretation remains stable when the aggregate benchmark is decomposed by source, scene, task category, fine-grained task type, input format, and response validity. We first check whether the results are dominated by a small number of scenes, then identify the ability categories and concrete task types that account for the main weaknesses, distinguish temporal reasoning difficulty from multi-image input complexity, and finally check whether answer-format failures explain the low scores. Unless otherwise specified, each analysis reports average accuracy across the 11 evaluated MLLMs.

\subsection{Robustness Across Sources and Scenes}
\label{app:source_specific_results}

\begin{figure}[!htbp]
  \centering
  \includegraphics[width=0.88\linewidth]{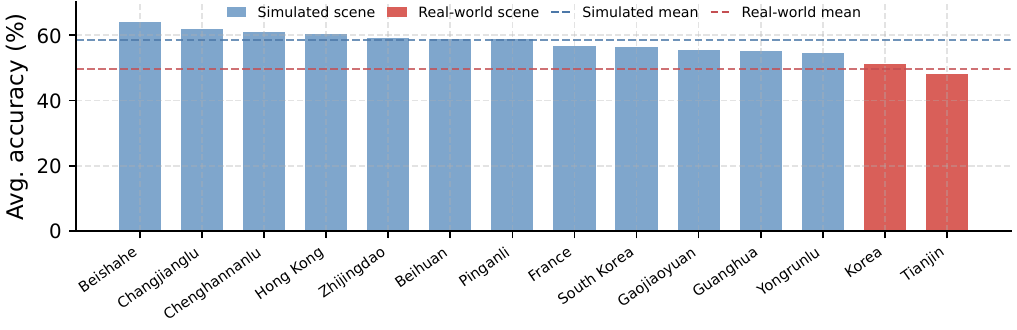}
  \caption{Scene-level average MLLM accuracy across simulated and real-world sources. Bar colors distinguish simulated reconstructed scenes from real-world roadside scenes; dashed lines mark the mean of each source.}
  \label{fig:app_scene_variation}
\end{figure}

We first ask whether the benchmark conclusions are an artifact of a small number of unusually easy or hard scenes. Fig.~\ref{fig:app_scene_variation} summarizes the average MLLM accuracy for each scene. In the simulated split, scene-level accuracy ranges from 54.6\% on Beijing Yongrunlu to 64.0\% on Beijing Beishahe, indicating moderate variation across reconstructed intersections rather than dependence on a single scene. The two real-world scenes are lower, with 51.2\% on Korea and 48.1\% on Tianjin. This source-level pattern suggests that benchmark difficulty is broadly distributed, while real-world roadside footage introduces additional challenges beyond the reconstructed simulation scenes.

Table~\ref{tab:app_scene_variation} reports the corresponding model-by-scene scores. The table shows that the trend in Fig.~\ref{fig:app_scene_variation} is not driven by a single model family: most models exhibit moderate variation across simulated scenes, while the real-world columns remain relatively difficult for both proprietary and open-source/open-weight models. Gemini-3-Pro is the strongest model on all simulated scenes and on Korea, yet its accuracy still drops from 67.8--79.0\% on simulated scenes to 60.7\% on Korea and 54.5\% on Tianjin. Other strong models show a similar source shift, with real-world scores generally concentrated around 48--55\%. These per-model results are consistent with the additional visual and structural complexity of roadside footage, including uncontrolled viewpoints, illumination and weather variation, occlusion, compression artifacts, dense markings, and heterogeneous road layouts.

\begin{table}[!htbp]
\centering
\scriptsize
\caption{Per-model scene-level accuracy on simulated and real-world sources (\%). Fig.~\ref{fig:app_scene_variation} visualizes scene-level trends after averaging over models; this table reports the underlying model-by-scene scores. The final row reports the scene average over the 11 evaluated MLLMs. Simulated scene abbreviations: BH=Beihuan, BS=Beishahe, CJ=Changjianglu, GJ=Gaojiaoyuan, PL=Pinganli, YR=Yongrunlu, CH=Chenghannanlu, GH=Guanghua, MS=Massy, HK=Yau Ma Tei, SG=Songdo, TJ=Zhijingdao.}
\label{tab:app_scene_variation}
\setlength{\tabcolsep}{3.2pt}
\resizebox{\linewidth}{!}{
\begin{tabular}{lrrrrrrrrrrrrrr}
\toprule
\textbf{Model} & \multicolumn{12}{c}{\textbf{Simulated scenes}} & \multicolumn{2}{c}{\textbf{Real-world}} \\
\cmidrule(lr){2-13}\cmidrule(lr){14-15}
& \textbf{BH} & \textbf{BS} & \textbf{CJ} & \textbf{GJ} & \textbf{PL} & \textbf{YR} & \textbf{CH} & \textbf{GH} & \textbf{MS} & \textbf{HK} & \textbf{SG} & \textbf{TJ} & \textbf{Korea} & \textbf{Tianjin} \\
\midrule
GPT-4o & 55.3 & 60.5 & 55.9 & 56.0 & 58.8 & 48.0 & 55.1 & 52.1 & 51.3 & 55.1 & 59.7 & 53.1 & 54.1 & \textbf{54.9} \\
GPT-5.2 & 57.4 & 62.2 & 61.8 & 52.3 & 58.8 & 57.8 & 62.9 & 55.1 & 58.9 & 55.6 & 59.2 & 56.1 & 48.4 & 48.9 \\
Gemini-2.5-Pro & 55.7 & 57.1 & 55.0 & 45.3 & 54.4 & 45.5 & 60.5 & 50.0 & 53.8 & 59.3 & 44.6 & 53.5 & 39.3 & 33.9 \\
Gemini-3-Pro & \textbf{75.0} & \textbf{75.2} & \textbf{73.2} & \textbf{73.7} & \textbf{77.2} & \textbf{73.8} & \textbf{74.2} & \textbf{77.5} & \textbf{75.8} & \textbf{79.0} & \textbf{67.8} & \textbf{69.3} & \textbf{60.7} & 54.5 \\
Claude-4.5 & 63.1 & 63.0 & 62.7 & 58.4 & 60.5 & 61.5 & 60.2 & 60.6 & 53.8 & 66.3 & 58.8 & 59.7 & 50.8 & 52.4 \\
Qwen-VL-Max & 62.7 & 67.2 & 62.3 & 54.3 & 57.9 & 57.8 & 62.1 & 57.2 & 60.2 & 59.1 & 57.1 & 64.9 & 51.6 & 49.8 \\
Qwen3-Plus & 54.9 & 63.0 & 59.1 & 52.7 & 53.5 & 46.7 & 64.5 & 46.6 & 58.1 & 55.1 & 59.2 & 58.3 & 49.2 & 41.2 \\
Qwen3-235B & 59.0 & 68.5 & 63.6 & 60.1 & 59.2 & 59.0 & 61.3 & 52.5 & 53.0 & 60.1 & 57.5 & 64.0 & 53.3 & 52.8 \\
Doubao-1.5 & 52.5 & 58.4 & 60.0 & 48.6 & 51.3 & 47.5 & 52.0 & 51.7 & 50.4 & 61.3 & 53.7 & 59.7 & 54.1 & 47.2 \\
InternVL & 53.7 & 64.7 & 58.6 & 49.0 & 53.5 & 44.3 & 56.2 & 45.8 & 48.3 & 51.9 & 46.8 & 57.5 & 46.7 & 40.3 \\
Grok-4 & 57.8 & 63.9 & 67.3 & 61.3 & 61.8 & 58.6 & 62.5 & 58.9 & 59.8 & 61.7 & 56.7 & 55.7 & 54.9 & 53.7 \\
\midrule
\textbf{Scene avg.} & \textbf{58.8} & \textbf{64.0} & \textbf{61.8} & \textbf{55.6} & \textbf{58.8} & \textbf{54.6} & \textbf{61.0} & \textbf{55.3} & \textbf{56.7} & \textbf{60.4} & \textbf{56.5} & \textbf{59.3} & \textbf{51.2} & \textbf{48.1} \\
\bottomrule
\end{tabular}
}
\end{table}

\subsection{Category and Fine-Grained Task Bottlenecks}
\label{app:category_task_bottlenecks}

\begin{figure}[!htbp]
  \centering
  \includegraphics[width=0.98\linewidth]{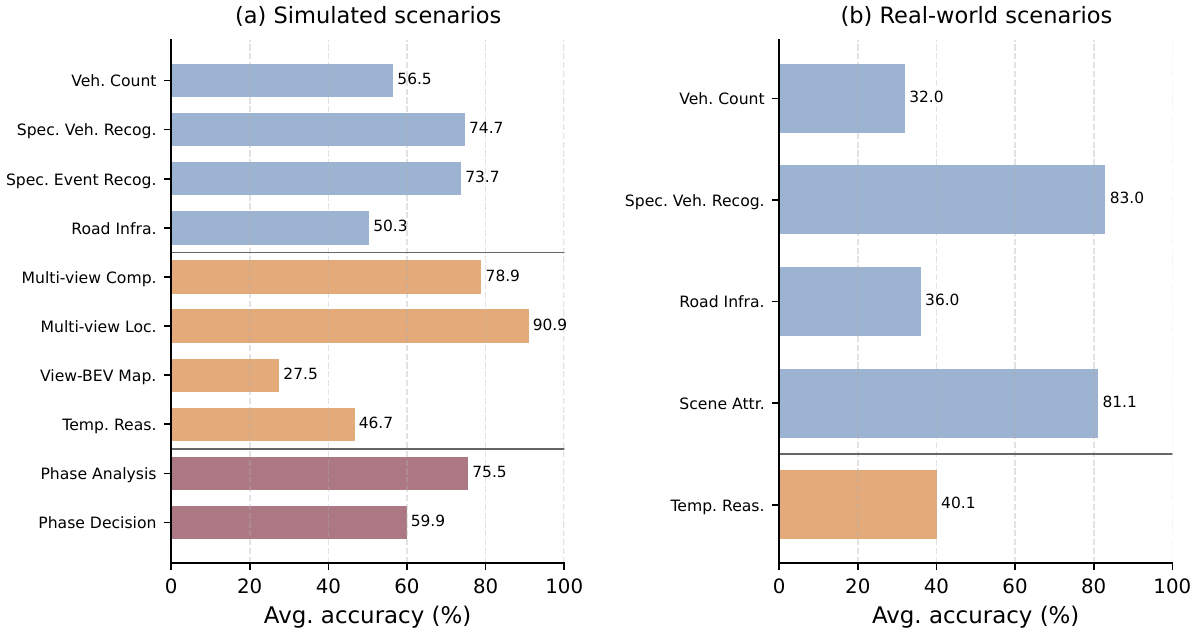}
  \caption{Category-level analysis on simulated and real-world scenarios. Bars report average accuracy across the 11 evaluated MLLMs. The real-world panel includes only categories supported by fixed roadside footage; multi-view and L3 phase-decision categories are simulation-only.}
  \label{fig:app_category_diagnostics}
\end{figure}

The scene-level results above establish that benchmark difficulty is not concentrated in a single scene, but they still average over heterogeneous task categories. Fig.~\ref{fig:app_category_diagnostics} therefore shifts the analysis from where errors occur to what kind of traffic understanding breaks down. In the simulated split, Multi-view Localization reaches 90.9\%, because a special vehicle often provides a salient visual anchor. However, View-BEV Mapping drops to 27.5\%, showing that successful cross-image matching does not imply geometric understanding of the intersection layout. Temporal Reasoning is also low at 46.7\%, indicating that frame ordering and queue evolution remain difficult even when the inputs are generated from controlled simulation metadata. In the real-world split, the supported categories show a different but consistent pattern: Special Vehicle Recognition and Scene Attribute remain relatively high, while Vehicle Counting, Road Infrastructure, and Temporal Reasoning are substantially lower. Thus, the category-level view clarifies the nature of the benchmark difficulty: model failures are not uniform visual degradation across all categories, but selective drops on tasks that require lane-level topology, quantitative traffic-state estimation, or temporal change reasoning.

\begin{figure}[!htbp]
  \centering
  \includegraphics[width=0.98\linewidth]{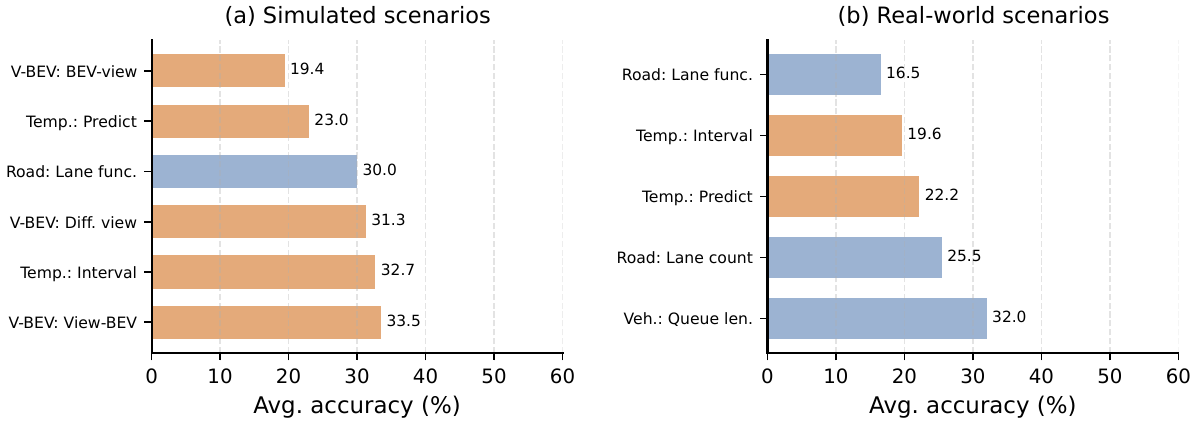}
  \caption{Lowest-scoring fine-grained tasks in the simulated and real-world splits. Bars report average accuracy across the 11 evaluated MLLMs.}
  \label{fig:app_task_bottlenecks}
\end{figure}

To make the category-level diagnosis more concrete, we further inspect the fine-grained task types under each category. Fig.~\ref{fig:app_task_bottlenecks} ranks the lowest-scoring tasks in each source. The lowest simulated tasks concentrate on View-BEV Mapping and Temporal Reasoning, where models must align camera views with a top-down layout or infer traffic evolution across time. The lowest real-world tasks instead emphasize Road Infrastructure and Temporal Reasoning, reflecting the added difficulty of lane-function recognition and temporal judgment under uncontrolled roadside imagery.

\subsection{Input Format and Response Validity}
\label{app:response_reliability}

The task-level analysis points repeatedly to temporal reasoning, but this weakness could still be confounded with input complexity: cross-timestep questions often contain multiple images. Table~\ref{tab:app_task_format_effects} therefore compares single-image, multi-image, and cross-timestep multi-image questions directly. In the simulated split, multi-image questions average 77.6\%, while cross-timestep multi-image questions fall to 38.7\%. This gap suggests that the difficulty is not simply the number of images: models can use multiple images when the answer is anchored by a salient object, but they struggle when the images must be ordered or extrapolated over time. The real-world split shows the same pattern. Cross-timestep accuracy is 40.1\%, below single-image accuracy at 53.8\%, even though real multi-image questions are limited to supported fixed-camera settings. These results strengthen the previous analyses by showing that temporal demand, rather than multi-image input alone, is a central source of difficulty.

\begin{table}[!htbp]
\centering
\small
\caption{Task-format effects on average MLLM accuracy. The last two columns report the cross-timestep multi-image accuracy minus the corresponding non-temporal format accuracy.}
\label{tab:app_task_format_effects}
\begin{tabular}{lccccc}
\toprule
\textbf{Source} & \textbf{Single image} & \textbf{Multi image} & \textbf{Cross-time} & \textbf{$\Delta$ vs. single} & \textbf{$\Delta$ vs. multi} \\
\midrule
Simulated & 60.4 & 77.6 & 38.7 & -21.7 & -38.9 \\
Real-world & 53.8 & 87.9 & 40.1 & -13.7 & -47.8 \\
\bottomrule
\end{tabular}
\end{table}

We also check whether the preceding failures could be explained by answer-format errors rather than traffic-scene misunderstanding. Table~\ref{tab:app_invalid_response_rates} reports the invalid-response rate after the same three-attempt retry protocol used in the main evaluation. Most models have near-zero invalid-response rates, so the low scores on View-BEV Mapping, Road Infrastructure, Vehicle Counting, and Temporal Reasoning are unlikely to be explained primarily by formatting failures. Gemini-2.5-Pro and InternVL are exceptions, with invalid-response rates of 15.8\% and 15.6\% on simulated scenarios and 33.0\% and 23.4\% on real-world scenarios. These models expose a second practical failure mode: constrained-answer compliance can degrade independently from traffic-scene understanding. A category-level invalid-response breakdown would further refine this check; here, the overall invalid-response rate serves as a conservative sanity check that the dominant benchmark weaknesses are reasoning-oriented rather than format-only artifacts.

\begin{table}[!htbp]
\centering
\small
\caption{Overall invalid-response rates after up to three inference attempts. Lower is better.}
\label{tab:app_invalid_response_rates}
\begin{tabular}{lcc}
\toprule
\textbf{Model} & \textbf{Simulated invalid (\%)} & \textbf{Real-world invalid (\%)} \\
\midrule
GPT-4o & 0.0 & 0.0 \\
GPT-5.2 & 0.1 & 0.0 \\
Gemini-2.5-Pro & 15.8 & 33.0 \\
Gemini-3-Pro & 0.0 & 0.0 \\
Claude-Sonnet-4.5 & 0.0 & 0.0 \\
Qwen-VL-Max & 0.0 & 0.0 \\
Qwen3-VL-Plus & 0.0 & 0.0 \\
Qwen3-VL-235B & 0.0 & 0.0 \\
Doubao-1.5 & 0.0 & 0.0 \\
InternVL & 15.6 & 23.4 \\
Grok-4 & 0.0 & 0.0 \\
\bottomrule
\end{tabular}
\end{table}

\paragraph{Summary.}
Together, these analyses turn the appendix into a validation of the main benchmark interpretation. The scene analysis shows that results are not dominated by a single source; the category analysis identifies the abilities behind the difficulty; the task-level analysis localizes those abilities to concrete traffic operations; the input-format analysis shows that temporal demand is harder than multi-image input alone; and the reliability analysis indicates that these weaknesses are not mainly answer-format artifacts. Across all views, model failures concentrate on topology-grounded perception, cross-view geometric alignment, queue-state estimation, and temporal reasoning.

\section{Detailed Sim-to-Real Transfer Analysis}
\label{app:sim2real_detail}

\subsection{Fine-Tuning Protocol}
\label{app:sim2real_finetuning_protocol}

This appendix summarizes the sim-to-real experiment behind Table~\ref{tab:results_single_column}. The real-world evaluation covers single-image perception, traffic-light state recognition, and cross-timestep temporal reasoning; synchronized multi-view phase-decision annotations are not available in the real-world footage, so L3 phase-level transfer is left for future evaluation. We use \textbf{Qwen3-VL-2B-Instruct} as the backbone model, first evaluate the base model on the real-world VQA benchmark, and then fine-tune the same model using only simulated OmniTraffic VQA samples. No real-world samples or human-verified benchmark items are used during fine-tuning.

For adaptation, we apply LoRA to attention-related modules in both visual and language branches, specifically \texttt{qkv} and \texttt{proj} in the vision stack and \texttt{q\_proj}, \texttt{k\_proj}, and \texttt{v\_proj} in the language stack. We do not adapt MLP/FFN blocks, as this design empirically provides a better balance between transfer effectiveness and overfitting risk under limited fine-tuning data. The LoRA hyperparameters are rank $r=16$, $\alpha=32$, and dropout $0.05$. Training uses an AdamW-style optimizer (via the default Hugging Face Trainer setup), with learning rate $2\times10^{-4}$, weight decay $0.01$, warmup ratio $0.03$, and a cosine learning-rate schedule. We set the micro-batch size to 1 and the gradient accumulation steps to 16, yielding an effective batch size of 16 samples per optimizer step. The maximum sequence length is set to 4096 to support mixed single-image and multi-image VQA inputs. All runs are conducted on a single A100 GPU.

\subsection{Transfer Results}
\label{app:transfer_results}

We evaluate transfer by comparing the fine-tuned model with the base model under the same real-world VQA protocol and deterministic decoding configuration. As shown in Table~\ref{tab:results_single_column}, simulated-only fine-tuning improves Qwen3-2B from 47.6\% to 53.8\% overall accuracy on real-world scenes, with a larger gain on Korea (+13.2 points) and a smaller gain on Tianjin (+2.6 points). This indicates that OmniTraffic simulation data provides useful supervision beyond the simulated domain.

\begin{table}[!htbp]
\centering
\small
\caption{Category-level effect of fine-tuning on OmniTraffic simulation data. Accuracy is reported before and after fine-tuning Qwen3-2B using only simulated OmniTraffic samples.}
\label{tab:app_sim2real_category_transfer}
\begin{tabular}{llccc}
\toprule
\textbf{Category} & \textbf{Coverage} & \textbf{Before} & \textbf{After} & \textbf{$\Delta$} \\
\midrule
Temporal Reasoning & Sim./Real & 34.0 & \textbf{50.0} & +16.0 \\
Special Vehicle Recognition & Sim./Real & 63.6 & \textbf{90.9} & +27.3 \\
Road Infrastructure & Sim./Real & 43.0 & \textbf{47.7} & +4.7 \\
Vehicle Counting & Sim./Real & \textbf{62.7} & 51.0 & -11.8 \\
Scene Attribute & Real-only & \textbf{75.0} & 65.0 & -10.0 \\
Traffic-Light State & Real-only & 70.6 & \textbf{76.5} & +5.9 \\
\bottomrule
\end{tabular}
\end{table}

Table~\ref{tab:app_sim2real_category_transfer} shows that the gains are concentrated in categories whose semantics are shared by simulation and reality. Temporal Reasoning and Special Vehicle Recognition improve most, suggesting that simulated supervision helps the model learn transferable temporal structure and traffic-specific semantic cues. Road Infrastructure also improves moderately, while Vehicle Counting and Scene Attribute drop, indicating that simulated-only fine-tuning can still hurt appearance-sensitive or real-only recognition under occlusion, viewpoint changes, and weather variation. Thus, simulation-only adaptation is useful for traffic-structure reasoning, but broader visual-domain coverage or limited real-world adaptation is still needed for robust perception.

\section{Qualitative Case Studies}
\label{app:qualitative_case_studies}

This section presents representative qualitative cases that complement the quantitative results. Rather than exhaustively cataloging all failure modes, we select one case from each cognitive level to illustrate how different models use visual evidence, track traffic states, and ground their answers in intersection structure.

\subsection{L1: Topology-Grounded Perception}
\label{app:qual_l1_topology_perception}

\begin{figure}[!htbp]
    \centering
    \includegraphics[width=0.66\linewidth]{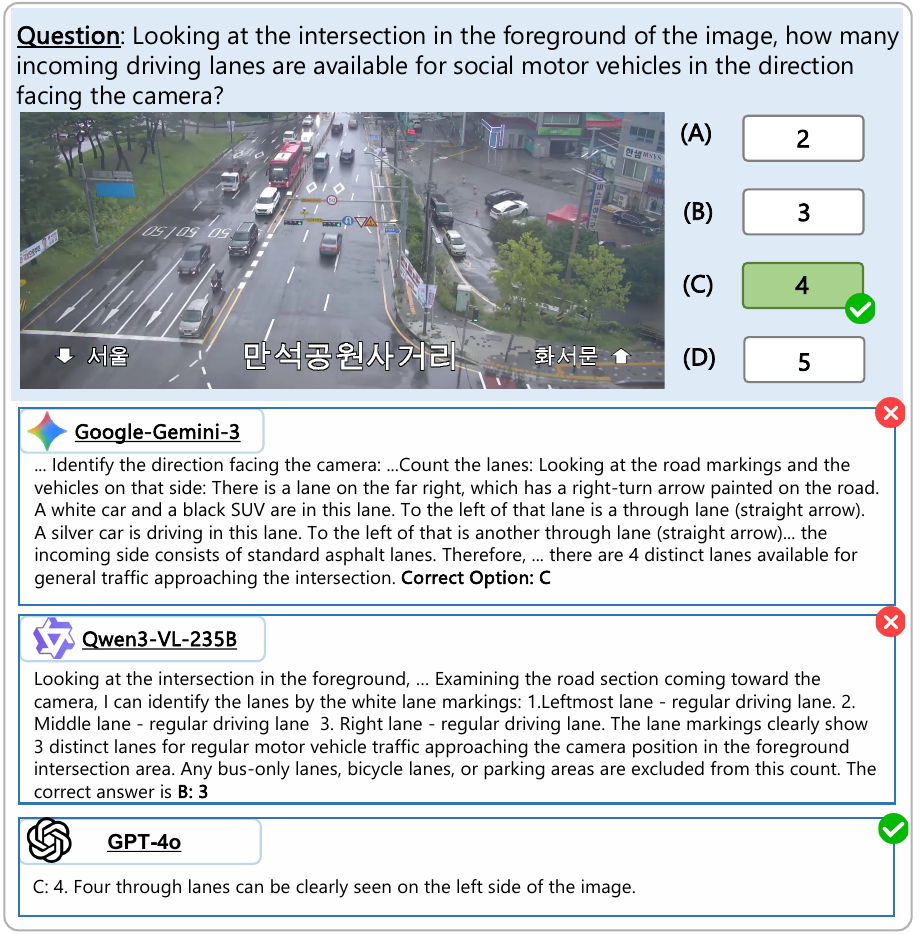}
    \caption{Representative L1 topology-grounded perception case. GPT-4o correctly identifies four incoming driving lanes for social motor vehicles, whereas Gemini-3-Pro and Qwen3-VL-235B miscount the lane structure in the real-world foreground intersection.}
    \label{fig:qual_l1_lane_count_case}
\end{figure}

Fig.~\ref{fig:qual_l1_lane_count_case} presents a representative L1 perception case from a real-world intersection. The question asks the model to count the number of incoming driving lanes available for social motor vehicles in the direction facing the camera. Among the three evaluated models, GPT-4o correctly answers four lanes, while Gemini-3-Pro and Qwen3-VL-235B provide incorrect counts.

This case illustrates that L1 perception in OmniTraffic requires more than generic road or vehicle recognition. The model must first determine the traffic direction specified by the question, separate social motor-vehicle lanes from other road regions, and count lane availability under real-world camera perspective and visual clutter. The incorrect responses show that even when models describe plausible lane markings and vehicle positions, they may still fail to ground those observations in the correct incoming direction and lane category. This supports the quantitative finding that real-world road-infrastructure perception remains challenging for current MLLMs.

\subsection{L2: Spatiotemporal Reasoning}
\label{app:qual_l2_spatiotemporal}

\begin{figure}[!htbp]
    \centering
    \includegraphics[width=0.66\linewidth]{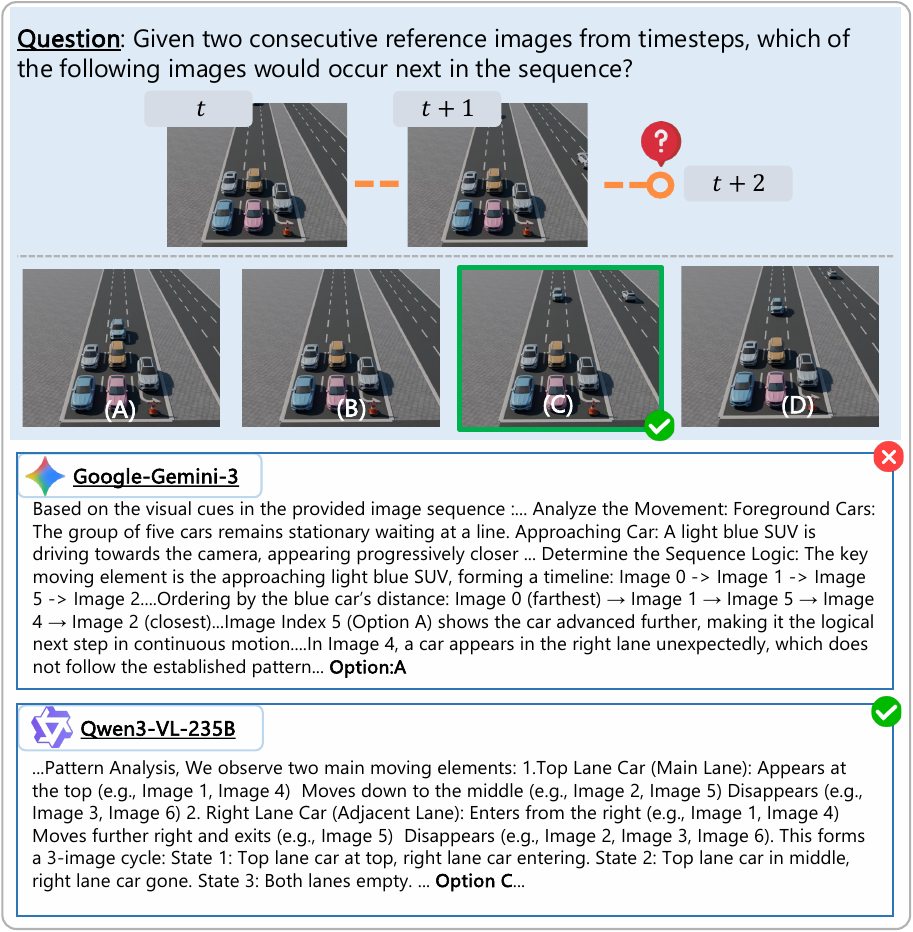}
    \caption{Representative L2 temporal reasoning case. Gemini-3-Pro incorrectly follows the most visually salient vehicle trajectory, whereas Qwen3-VL-235B correctly reasons over the joint temporal pattern of multiple vehicles across lanes.}
    \label{fig:qual_l2_temporal_case}
\end{figure}

Fig.~\ref{fig:qual_l2_temporal_case} presents a representative temporal reasoning case. Given two consecutive reference frames, the model is asked to infer the next frame from multiple candidates. This example contrasts two different reasoning behaviors: Gemini-3-Pro selects an incorrect option by focusing on the apparent motion of a salient vehicle, while Qwen3-VL-235B correctly identifies the underlying multi-object state transition across lanes.

This case illustrates that L2 failures are not merely caused by object-recognition errors. The incorrect answer still reflects plausible local motion tracking, but it misses the structured evolution of the traffic state. Correct temporal reasoning requires models to compare multiple vehicles jointly, track their lane-specific state changes, and infer the next frame from the overall scene dynamics rather than from a single salient object.

\subsection{L3: Decision Support}
\label{app:qual_l3_decision_support}

\begin{figure}[!htbp]
    \centering
    \includegraphics[width=0.66\linewidth]{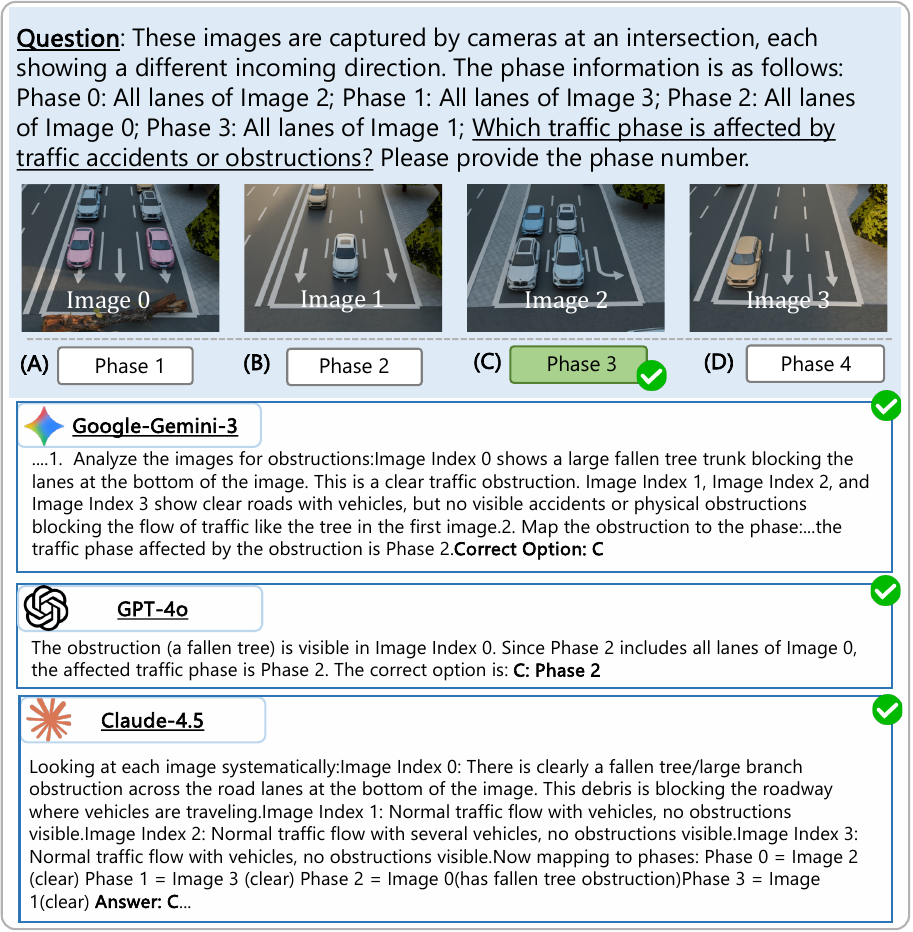}
    \caption{Representative L3 decision-support case. All three models correctly identify the obstruction in Image~0 and map it to Phase~2, showing that salient physical obstructions can be grounded in the provided image-to-phase correspondence.}
    \label{fig:qual_l3_phase_obstruction_case}
\end{figure}

Fig.~\ref{fig:qual_l3_phase_obstruction_case} presents a representative L3 decision-support case. The input contains four incoming-direction views from the same intersection, together with a phase definition that maps each phase to all lanes in one image. The model is asked to identify which traffic phase is affected by an accident or obstruction. Gemini-3-Pro, GPT-4o, and Claude-Sonnet-4.5 all correctly locate the fallen-tree obstruction in Image~0 and map it to Phase~2 according to the provided phase definition.

This case shows that current MLLMs can combine local visual grounding with symbolic phase mapping. Unlike the L2 temporal case, the key evidence is spatially salient, and the phase definition explicitly links images to traffic phases. The successful responses suggest that MLLMs can provide reliable decision-support reasoning when the traffic event is visually distinctive and the required mapping is directly specified in the prompt.

\section{Discussion}
\label{app:discussion}

The results of OmniTraffic highlight a gap between general-purpose multimodal understanding and deployable traffic intelligence. The main bottleneck is not generic visual recognition, but structure-aware spatiotemporal reasoning. In traffic scenes, objects must be interpreted through the operational geometry of an intersection: vehicles are associated with lanes and queues, camera views must be aligned with BEV layouts, temporal changes must be tracked, and signal phases must be grounded in observed traffic states. Current MLLMs remain weak in these connected capabilities, which limits the reliability of their phase-level decisions and exposes the need for traffic-specific evaluation beyond conventional visual QA.

OmniTraffic further shows that simulation-generated supervision can help narrow this gap. Real-world traffic data is difficult to scale because synchronized multi-view footage is rare, topology-level and phase-level annotations are expensive, and safety-critical events are hard to collect repeatedly. By contrast, simulation provides controllable traffic states, precise metadata, synchronized views, and configurable rare events. Our sim-to-real results demonstrate that fine-tuning on simulated OmniTraffic samples improves real-world traffic understanding, indicating that simulation is not merely a data substitute but a practical route for strengthening traffic-specific cognition in MLLMs. Nevertheless, the current transfer study is still limited by simulation realism and by the relatively simple simulated-only supervised fine-tuning recipe; more realistic rendering of weather, illumination, occlusion, and camera artifacts, together with mixed simulated-real adaptation or more carefully tuned SFT strategies, may reduce negative transfer on appearance-sensitive categories.

This is why OmniTraffic is designed as an open pipeline rather than a closed dataset. Beyond releasing images and benchmark questions, we release reconstructed 3D intersection assets, traffic metadata, rendering scripts, and the VQA generation pipeline. These resources allow users to customize camera placements, traffic demand profiles, rare events, visual conditions, and scene configurations without rebuilding geometry or annotation logic from scratch. In this sense, OmniTraffic not only identifies current model limitations and validates a simulation-based improvement path, but also provides infrastructure for extending traffic-oriented multimodal intelligence.

\end{document}